\PassOptionsToPackage{table}{xcolor}

\documentclass[10pt,twocolumn,letterpaper]{article}

\usepackage[pagenumbers]{cvpr} 
\usepackage{multirow}
\usepackage{comment}

\usepackage[dvipsnames]{xcolor}

\definecolor{cvprblue}{rgb}{0.21,0.49,0.74}
\usepackage[pagebackref,breaklinks,colorlinks,citecolor=cvprblue]{hyperref}
\usepackage[table]{xcolor}
\usepackage{bm}
\usepackage{threeparttable}
\usepackage{multirow}
\usepackage{graphicx}
\usepackage{multicol}
\usepackage{comment}

\usepackage{amsmath}
\usepackage{algorithm}
\usepackage{algpseudocode}

\newcommand{\vect}[1]{\mathbf{#1}}

\usepackage[table]{xcolor}
\usepackage{booktabs}
\usepackage{siunitx}
\usepackage{tikz}
\usepackage{makecell} 
\usepackage{multirow} 

\usepackage{amsmath}
\usepackage{amssymb}
\usepackage{cuted}
\usepackage{gensymb}
\usepackage{tabularx}
\usepackage{array}
\usepackage{tablefootnote}
\usepackage{listings}

\definecolor{codegreen}{rgb}{0,0.6,0}
\definecolor{codegray}{rgb}{0.5,0.5,0.5}
\definecolor{codepurple}{rgb}{0.58,0,0.82}
\definecolor{backcolour}{rgb}{0.95,0.95,0.92}
\lstdefinestyle{mystyle}{
    backgroundcolor=\color{backcolour},   
    commentstyle=\color{codegreen},
    keywordstyle=\color{magenta},
    stringstyle=\color{codepurple},
    basicstyle=\ttfamily\footnotesize,
    breakatwhitespace=false,         
    breaklines=true,                 
    captionpos=b,                    
    keepspaces=true,                 
    showspaces=false,                
    showstringspaces=false,
    showtabs=false,                  
    tabsize=2
}
\author{
Soonbin Lee$^{1}$ \and
Yeong-Gyu Kim$^{2}$ \and
Simon Sasse$^{1}$ \and
Tomás M. Borges$^{1}$ \and
Yago Sánchez$^{1}$ \and
Eun-Seok Ryu$^{2}$ \and
Thomas Schierl$^{1}$ \and
Cornelius Hellge$^{1}$ \and
$^{1}$Fraunhofer Heinrich-Hertz-Institute (HHI), Germany \\
$^{2}$Sungkyunkwan University (SKKU), South Korea \\
{\tt\small \{first\_name\}.\{last\_name\}@hhi.fraunhofer.de},
{\tt\small \{kyk2798, esryu\}@skku.edu}
}

\title{GaussianPOP: Principled Simplification Framework for \\Compact 3D Gaussian Splatting via Error Quantification}

\begin{document}

\twocolumn[{
\maketitle
\vspace{-4.5mm}
\centering
    \captionsetup{type=figure}
    \includegraphics[width=\textwidth]{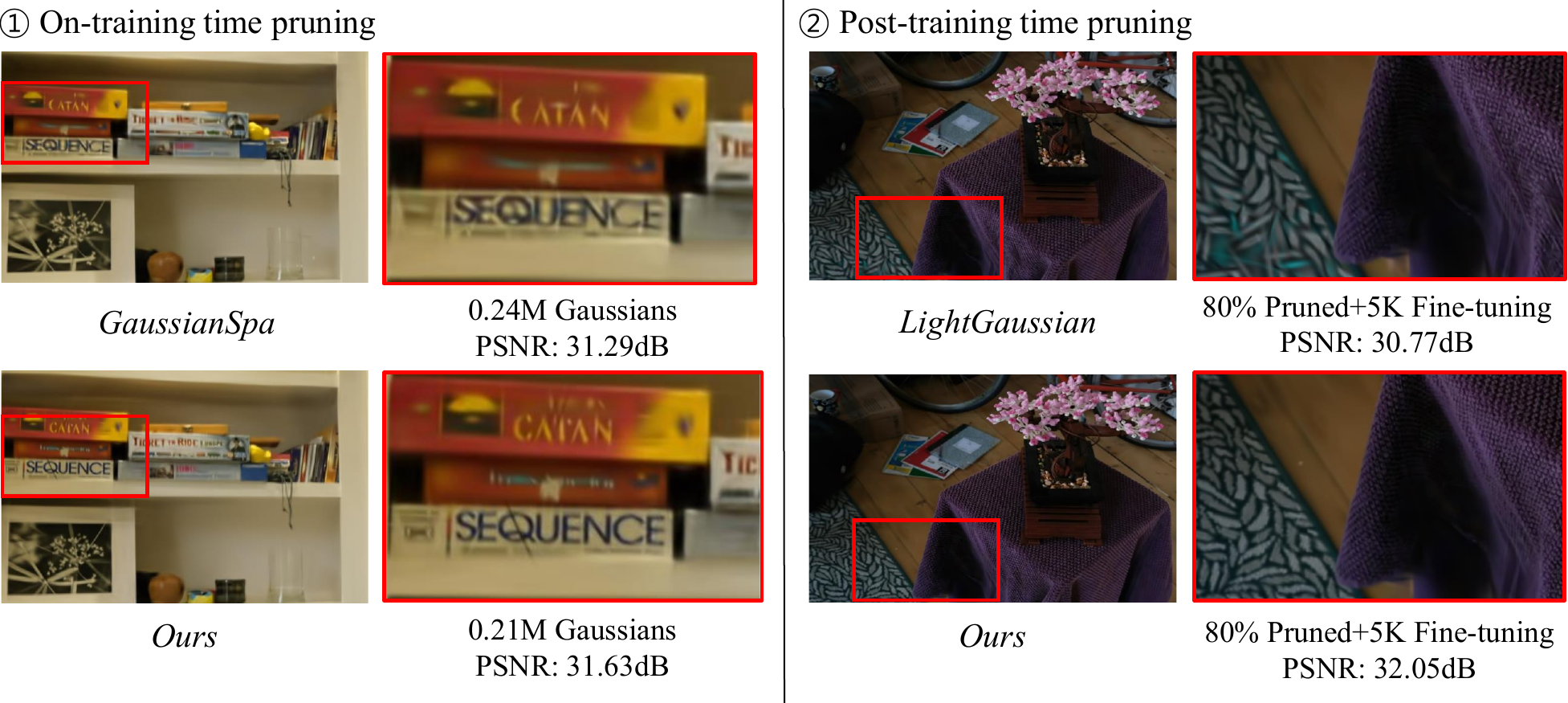}
    \captionof{figure}{We propose \texttt{\textbf{GaussianPOP}}, enabling high-quality and compact view synthesis with superior rendering of details. The figure compares our method against state-of-the-art approaches in two key scenarios. The on-training scenario integrates pruning into the training process from scratch, whereas the post-training scenario applies simplification to a pre-trained model. 
    }
    \label{fig:fig1}
    \vspace{5.0mm}
}]

\begin{abstract}
Existing 3D Gaussian Splatting simplification methods commonly use importance scores, such as blending weights or sensitivity, to identify redundant Gaussians. However, these scores are not driven by visual error metrics, often leading to suboptimal trade-offs between compactness and rendering fidelity. We present GaussianPOP, a principled simplification framework based on analytical Gaussian error quantification. Our key contribution is a novel error criterion, derived directly from the 3DGS rendering equation, that precisely measures each Gaussian’s contribution to the rendered image. By introducing a highly efficient algorithm, our framework enables practical error calculation in a single forward pass. The framework is both accurate and flexible, supporting on-training pruning as well as post-training simplification via iterative error re-quantification for improved stability. Experimental results show that our method consistently outperforms existing state-of-the-art pruning methods across both application scenarios, achieving a superior trade-off between model compactness and high rendering quality.  
\end{abstract}

\vspace{-4.5mm}

\section{Introduction}
\label{sec:intro}

\begin{table}[t]
\begingroup 
\renewcommand{\arraystretch}{1.1} 
\centering
\small
\begin{tabularx}{\columnwidth}{l >{\centering\arraybackslash}X} 
\toprule
\textbf{Methods} & \textbf{Pruning Criteria} \\
\midrule
3DGS \cite{kerbl2023}& $\alpha\  (\text{opacity})$ \\
Compact3DGS \cite{lee2023compact}  & $L_{m}=\sum_{\alpha,s}(m)$ \\
MaskGaussian \cite{11094772} & $L_{m}=(\sum\mathcal{M}_{i})^{2}$ \\
GaussianSpa \cite{11092501}  & \makecell{$\min_{a, \Theta} \mathcal{L}(a, \Theta),$ $\text{s.t. } ||\boldsymbol{a}||_0 \le \kappa$} \\
C3DGS  \cite{niedermayr2023compressed}       & $S(p)=|\frac{\partial E}{\partial p}|$ \\
Mini-Splatting \cite{10.1007/978-3-031-72980-5_10} & $I_{i}=\sum w_{ij}$, $w_{ij}= T_{ij} \alpha_{ij}$ \\
LightGaussian \cite{fan2023lightgaussian}& $GS_{j}=\sum \mathbb{I}(G_j, r_i) \cdot \alpha_j \cdot T \cdot \gamma(\Sigma_j)$ \\
MesonGS \cite{10.1007/978-3-031-73414-4_25}  & $I_g = I_i \cdot (V_{\text{norm}})^{\beta}$ \\
PUP-3DGS \cite{11092397}     & $U_{i}=\log|\nabla_{x_{i},s_{i}}I_{\mathcal{G}}\nabla_{x_{i},s_{i}}I_{\mathcal{G}}^{T}|$ \\
\midrule
\rowcolor{black!10}
\textbf{Ours} & $\Delta SE_k = \|T_{k} \alpha_k \Delta c_k\|^2$ \\
\bottomrule
\end{tabularx}
\caption{Comparison of pruning criteria employed by various 3D Gaussian Splatting simplification methods.}
\label{tab:tab1}
\endgroup
\end{table}

3D Gaussian Splatting (3DGS) \cite{kerbl2023} achieves real-time, photorealistic novel view synthesis using millions of anisotropic 3D Gaussian primitives. Despite its impressive results, the resulting dense representation requires substantial memory and storage. Simplifying 3DGS models by reducing the number of Gaussians is a critical research area, as this count is the primary factor directly influencing model size, rendering performance, and deployment feasibility.

Current simplification approaches often evaluate a Gaussian for pruning using importance-based scores \cite{10.1007/978-3-031-72980-5_10, 10.1007/978-3-031-73414-4_25}. These methods typically employ rendering-based scores, such as accumulated blending or gradient sensitivity \cite{fan2023lightgaussian, ali2024elmgsenhancingmemorycomputation, 11094159}, to estimate the visual error rather than to quantify the error itself. Relying on these metrics that lack a direct correlation with visual impact often yields suboptimal trade-offs between compactness and fidelity. Therefore, a clear need exists for a simplification technique that is both principled in evaluating visual error and flexible in its application. We propose GaussianPOP, a simplification framework centered on an analytically-derived criterion that directly quantifies the visual error of removing a Gaussian. Through extensive experiments, we show that our method consistently outperforms existing state-of-the-art simplification techniques focused on reducing the number of Gaussians, achieving a superior trade-off between model compactness and rendering quality. Our contributions are as follows:
\begin{itemize}
    \item We derive a principled error criterion that analytically quantifies the pixel error from removing a single Gaussian, providing a more efficient and accurate pruning criterion.
    \item We introduce a highly efficient algorithm to compute this error for all Gaussians in a scene, integrated seamlessly into the 3DGS pruning process.
    \item We demonstrate the flexibility of our framework as shown in Fig. \ref{fig:fig1}, enabling it to function both as an on-training-time pruning schedule and as a post-training simplification tool.
\end{itemize}
\section{Related Work}
\label{sec:relatedwork}

\begin{figure*}[t]
 \centering 
 \includegraphics[width=\linewidth]{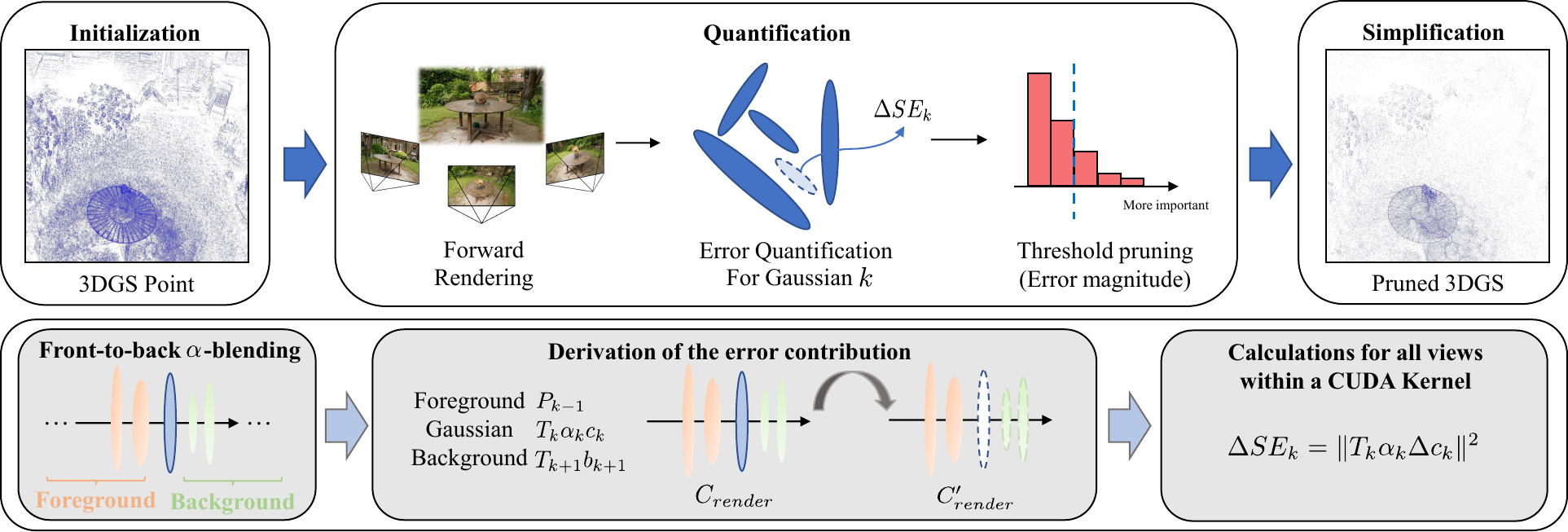}
 \caption{\textbf{Overview of the GaussianPOP simplification pipeline.} We quantify the per-pixel error ($\Delta SE_k$) for each Gaussian by analytically deriving its visual contribution from the $\alpha$-blending equation. This error is calculated efficiently, allowing us to prune low-error Gaussians to create a compact model.} 
 \label{fig:fig2}
\end{figure*}

Existing approaches to reduce the number of Gaussians in 3DGS models can be grouped into two categories. The first category, herein referred to as criterion-based pruning, evaluates individual Gaussians using importance-based metrics. Examples include opacity-based \cite{kerbl2023,mcmc, bogauss}, gradient-sensitivity (e.g, C3DGS \cite{niedermayr2023compressed} and PUP \cite{11092397}), or based on blending contribution and alpha-transmittance \cite{11125567, 10.1007/978-3-031-72980-5_10, 10.1007/978-3-031-73414-4_25, 10.1145/3680528.3687694, 11094159}. The key limitation of the approaches of this first category is their reliance on surrogate scores (e.g., Mini-Splatting \cite{10.1007/978-3-031-72980-5_10, fang2024minisplatting2building360scenes} and LightGS \cite{fan2023lightgaussian}), which are not perfect estimates of the visual quality impact when removing the associated Gaussians. Therefore, these criteria often overlook the actual visual error resulting from the removal of a Gaussian. For instance, a Gaussian with a large scale might seem important but could be completely occluded and thus visually irrelevant. In addition, a Gaussian with low opacity might seem insignificant, yet be critical for a necessary semi-transparent effect.

The second category, i.e., sparsity induction, jointly optimizes for simplification and reconstruction. This is achieved either through learnable masks \cite{lee2023compact, 10.5555/3737916.3741807} or by formulating the simplification as  constrained optimization problems, as can be seen in GaussianSpa \cite{11092501} and Maskgaussian \cite{11094772}. One limitation of this approach is its lack of generality. By fundamentally changing the training objective to encourage sparsity, these methods introduce a compromise: the model is pushed toward a sparse solution, which can prevent it from converging to the high-fidelity representation of the original 3DGS model. Furthermore, these methods are not general-purpose simplification tools, as they require a specialized training process from scratch and cannot be applied to pre-trained models.

Compression techniques also contribute to reducing the size of 3DGS representations. Distinct from methods that reduce the number of Gaussians, another line of work focuses on compressing the attributes \cite{3dgszip, entropygs, reduce10.1145/3651282}. These methods employ techniques such as vector quantization (VQ) for attributes \cite{wang2024rdogaussian, lee2024compact3ddynamic, lee2025omg, lvq}, parameter factorization \cite{10.1145/3664647.3681116, 10.1145/3664647.3681468, locogs}, or as seen in recent works predicting attributes from compact latent representations \cite{girish2023eagles, hac2024, lee2025compression3dgaussiansplatting, cat3dgs, ali2024elmgsenhancingmemorycomputation,sparsity}. Although these methods are effective for storage reduction and can be complementary to pruning, we consider compression of 3DGS attributes a distinct problem from simplification, as, for instance, it does not reduce the rendering complexity. Therefore, this paper focuses on the simplification of 3DGS specifically based on the reduction of the Gaussian count.

Tab. \ref{tab:tab1} summarizes the criteria these methods use. As a criterion-based approach, GaussianPOP benefits from the high flexibility of this category, allowing it to simplify pre-trained models. Moreover, our method is analytically derived to directly quantify visual error. As we will demonstrate, this combination of criterion-based flexibility and accuracy allows GaussianPOP to consistently outperform both existing criterion-based methods and the optimized sparsity-induction models.

\begin{algorithm}[h]
\caption{`Render-once' Efficient Error Computation (Per-pixel)}
\label{alg:error_quant}
\begin{algorithmic}[1]
\Statex $\triangleright$ \textbf{--- Phase 1: Compute list $H$ with forward rendering ---}
\State $\vect{C}_{render} \in \mathbb{R}^3$ (The final color of the rendered pixel)
\State $H = \langle (ID, \vect{c}_k, \alpha_k) \rangle_{k=1}^N$ (An ordered list of $N$ contributing Gaussians, sorted front-to-back, where $ID$ is a global identifier, $\vect{c}_k \in \mathbb{R}^3$ is the $k$-th base color, and $\alpha_k \in [0, 1]$ is the opacity)
\State $\Delta_{\text{SE}}$ (A global array, indexed by Gaussian $ID$, storing cumulative error)
\State $\mathcal{P}$ (A local array of size $N$ to store cumulative color sums)
\State $\mathcal{T}$ (A local array of size $N$ to store cumulative transmittances)
\Statex
\Statex $\triangleright$ \textbf{--- Phase 2: Compute Cumulative Color and Transmittance ---}
\State $\vect{P}_{\text{sum}} \gets 0$
\State $T \gets 1.0$
\For{$k \gets 1$ to $N$}
    \State $(ID, \vect{c}_k, \alpha_k) \gets H_k$
    \State $\vect{P}_{\text{sum}} \gets \vect{P}_{\text{sum}} + T \alpha_k  \vect{c}_k$
    \State $T^{\prime} \gets T (1 - \alpha_k)$
    \State $\mathcal{P}_k \gets \vect{P}_{\text{sum}}$ \Comment{Store $P_k = \sum_{j=1}^{k} (T_j \alpha_j \vect{c}_j)$}
    \State $\mathcal{T}_k \gets T^{\prime}$ \Comment{Store $T_{k} = \prod_{j=1}^{k-1} (1 - \alpha_j)$}
\EndFor
\Statex
\Statex $\triangleright$ \textbf{--- Phase 3: Compute Error Contribution based on Color Difference ---}
\For{$k \gets 1$ to $N$}
    \State $(ID, \vect{c}_k, \alpha_k) \gets H_k$
    \State $\vect{P}_k \gets \mathcal{P}_k$ \Comment{Cumulative color $\vect{P}_k$}
    \State $T_{k+1} \gets \mathcal{T}_k$ \Comment{Transmittance after $k$ ($T_{k+1}$)}
    \If{$k = 1$}
        \State $T_k \gets 1.0$ 
    \Else
        \State $T_k \gets \mathcal{T}_{k-1}$ \Comment{Transmittance before $k$ ($T_k$)}
    \EndIf
    \Statex \Comment{Analytically solve for $\vect{b}_{k+1}$ from the equation:~~~
    \Statex \hspace{3.2em}$\vect{C}_{render} = \vect{P}_k + (T_{k+1}\vect{b}_{k+1})$}
    \State \colorbox{white!100}{$\vect{b}_{k+1} \gets (\vect{C}_{render} - \vect{P}_k) / (T_{k+1} + \epsilon)$}
    \Statex \Comment{Calculate the Squared Error (SE) for $k$ (Eq. 4)~~~~}
    \State \colorbox{white!100}{$\Delta \mathbf{c}_k \gets \mathbf{c}_k - \mathbf{b}_{k+1}$}
    \State \colorbox{gray!25}{$\Delta SE_k \gets \| (T_k \alpha_k) \Delta \mathbf{c}_k \|^2$}
    \Statex \Comment{Accumulate $\Delta SE_{k}$ to the global Gaussian ID~~~~~}
    \State \colorbox{white!100}{$\Delta_{\text{SE}}[ID] \gets \Delta_{\text{SE}}[ID] + \Delta SE_k$}
\EndFor
\end{algorithmic}
\end{algorithm}

\section{Proposed Method}

Our method, illustrated in Fig.~\ref{fig:fig2}, introduces a principled error criterion derived directly from the 3DGS rendering equation. First, we revisit the 3DGS rendering process (Sec. \ref{sec:sec3.1}), then analytically derive the proposed error quantification (Sec. \ref{sec:sec3.2}), detail the efficient single-pass algorithm for its computation (Sec. \ref{sec:sec3.3}), and finally outline the complete simplification framework (Sec. \ref{sec:sec3.4}).

\subsection{Background: 3DGS Rendering}
\label{sec:sec3.1}
The 3DGS rendering process relies on projecting 3D Gaussians onto the image plane and performing an order-dependent alpha-blending. For a single pixel, the rendered color $C_{render}$ is the composite of all $N$ contributing Gaussians, sorted front-to-back:
\begin{equation}
C_{render} = \sum_{k=1}^{N} T_{k}\alpha_{k}c_{k}
\end{equation}
where $c_{k}$ is the base color of the $k$-th Gaussian, $\alpha_{k}$ is its pixel-specific opacity, and $T_{k} = \prod_{j=1}^{k-1}(1-\alpha_{j})$ is the transmittance. In our framework, $C_{render}$ denotes the final color rendered by the current set of Gaussians, serving as the baseline for error quantification. Therefore, the proposed method does not specifically require ground truth views in the quantification process. This, like other criterion-based methods, provides the flexibility to simplify any pre-trained 3DGS model.

\subsection{Principled Error Quantification}
\label{sec:sec3.2}
As discussed in Sec. \ref{sec:relatedwork}, existing simplification methods typically rely on importance-based metrics. These criteria often include accumulating the blending weight or employing gradient- and sensitivity-based metrics. Since these scores provide only an indirect measure of visual impact and may not perfectly reflect it, we formulate a criterion that serves as a direct approximation of the visual error itself. 

We analytically compute the squared error (SE) per-pixel that would be induced by removing a single Gaussian $k$ from the sorted list compared to using all Gaussians. Intuitively, this error is the squared difference between the original color $C_{render}$ and the new counterfactual color $C_{render}^{\prime}$ that would appear if $k$ was removed. To formalize this, we can express the original color $C_{render}$ as the sum of three parts: (1) the color accumulated before $k$ ($P_{k-1}$), (2) the contribution of $k$ itself $T_{k}\alpha_{k}c_{k}$, and (3) the contribution of the background after $k$ ($b_{k+1}$), which is attenuated by $k$'s presence $T_{k+1}b_{k+1}=T_{k}(1-\alpha_{k})b_{k+1}$,
\begin{equation}
C_{render} = \underbrace{P_{k-1}}_{\text{foreground}} + \underbrace{T_{k}\alpha_{k}c_{k}}_{\text{Gaussian } k} + \underbrace{T_{k}(1-\alpha_{k})b_{k+1}}_{\text{background}}
\end{equation}
If we remove Gaussian $k$, its contribution vanishes, and the background $b_{k+1}$ is no longer attenuated by $\alpha_{k}$. Thus, the counterfactual color $C_{render}^{\prime}$ is just the foreground blending directly with the background (see Fig.~\ref{fig:fig2}),
\begin{equation}
C_{render}^{\prime} = \underbrace{P_{k-1}}_{\text{foreground}} + \underbrace{T_{k}b_{k+1}}_{\text{background}}
\end{equation}
The per-pixel SE induced by the removal of Gaussian $k$ is $\Delta SE_{k} = ||C_{render} - C_{render}^{\prime}||^{2}$. Substituting the two expressions above, the foreground terms $P_{k-1}$ cancel out, and this SE calculation simplifies to the difference between the Gaussian's color $c_k$ and the background color $b_{k+1}$, weighted by the Gaussian's effective contribution $T_{k}\alpha_{k}$,
\begin{equation}
\left\|C_{render}-C_{render }^{\prime}\right\|^2=\left\|T_k \alpha_k\left(c_k-b_{k+1}\right)\right\|^2
\end{equation}

\begin{figure}[t]
 \centering
      \begin{subfigure}[b]{\linewidth}
         \centering
         \includegraphics[width=\linewidth]{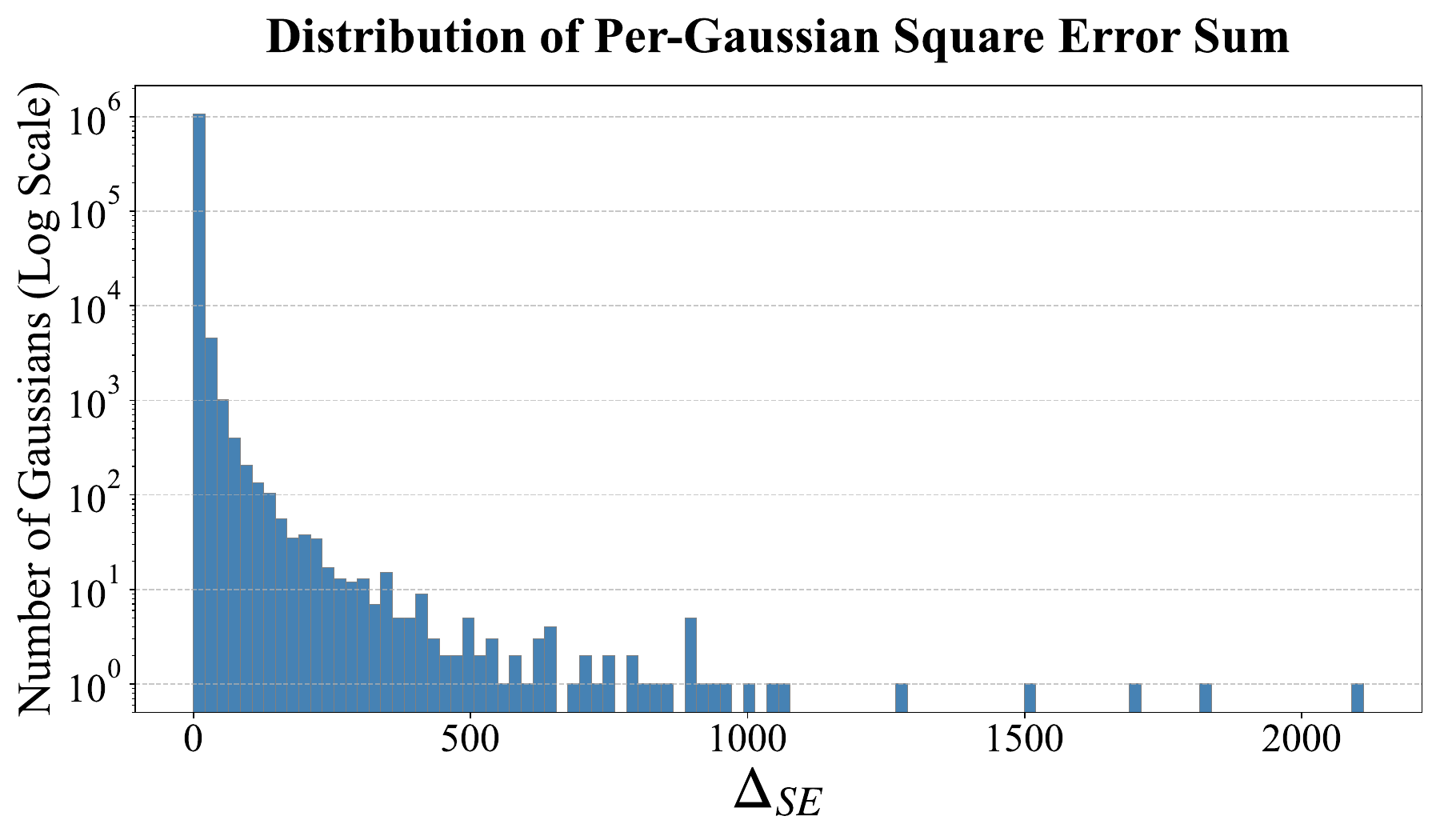}
     \end{subfigure}
 \caption{\textbf{Distribution of the cumulative square error $\Delta_{SE}$ across all training views for pre-trained `Bonsai' scene.} The majority of Gaussians contribute near-zero error, indicating that most Gaussians are non-essential to the final render.}
 \label{fig:fig3}
\end{figure}

However, quantifying the error contribution of each Gaussian presents a significant computational challenge. A naive approach would require iteratively removing each Gaussian, one at a time, and re-rendering all training views to measure the resulting error. For a scene with $M$ Gaussians and $V$ views, this process necessitates $M \times V$ full rendering passes, which is computationally prohibitive. This bottleneck arises because every rendering pass necessitates re-fetching the entire set of $M$ Gaussians from VRAM, which makes naive quantification computationally intractable.

\subsection{Practical Computation of the Error Criterion}
\label{sec:sec3.3}
We designed our framework specifically to circumvent this overhead. In contrast to the naive approach requiring $M$ expensive re-rendering operations, our approach uses a `render-once, compute-locally' strategy, as detailed in Algorithm~\ref{alg:error_quant}. This strategy begins with Phase 1, a standard forward rendering pass to generate the ordered list $H$ of $N$ contributing Gaussians. The algorithm then proceeds in two distinct computational stages. In Phase 2, a parallel prefix sum operation is executed over $H$. This pass efficiently computes the final rendered pixel color $C_{render}$ and simultaneously caches the cumulative color $\mathcal{P}_k$ and cumulative transmittance $\mathcal{T}_k$ values. In Phase 3, the kernel leverages this local cache. This stage is highly parallelizable, as the error computation for each Gaussian $k$ is independent. It analytically derives the background color $\vect{b}_{k+1}$ using only the final $C_{render}$ and the cached values $\mathcal{P}_k, \mathcal{T}_k$. Here $\epsilon=1e^{-9}$ is for numerical stability. This enables the $\Delta SE_k$ calculation using only values in local registers, which are then atomically accumulated. This strategy substitutes the intractable re-renders required by a naive approach with computationally parallel local operations. Consequently, the quantification stage is highly efficient, making it practical for both on training-time and post-training simplification.

\newlength{\textgaugelength}
\newlength{\barwidth}

\newcommand{\gmaxM}{3.110} 
\newcommand{\gminM}{0.373} 
\newcommand{\gmaxT}{1.830} 
\newcommand{\gminT}{0.180} 
\newcommand{\gmaxD}{2.780} 
\newcommand{\gminD}{0.208} 

\newcommand{\pmaxM}{27.53} 
\newcommand{\pminM}{26.81} 
\newcommand{\pmaxT}{23.73} 
\newcommand{\pminT}{23.10} 
\newcommand{\pmaxD}{30.22} 
\newcommand{\pminD}{27.57} 

\newcommand{\makegaugeG}[4]{%
  \settowidth{\textgaugelength}{#4}%
  \pgfmathsetmacro{\percent}{(0.6 / (#2 - #3)) * (#1 - #2) + 1.0}%
  \pgfmathsetmacro{\percent}{min(max(\percent, 0.3), 1.0)}%
  \setlength{\barwidth}{\percent\textgaugelength}%
  \rlap{\makebox[\textgaugelength][r]{\color{green!30}\rule{\barwidth}{0.8em}}}%
  #4%
}

\newcommand{\makegaugeP}[4]{%
  \settowidth{\textgaugelength}{#4}%
  \pgfmathsetmacro{\percent}{(0.6 / (#2 - #3)) * (#1 - #3) + 0.4}%
  \pgfmathsetmacro{\percent}{min(max(\percent, 0.7), 1.0)}%
  \setlength{\barwidth}{\percent\textgaugelength}%
  \rlap{\makebox[\textgaugelength][l]{\color{red!20}\rule{\barwidth}{0.8em}}}%
  #4%
}

\newcommand{\gaugebarGM}[1]{\makegaugeG{#1}{\gmaxM}{\gminM}{#1}}
\newcommand{\gaugebarboldGM}[1]{\makegaugeG{#1}{\gmaxM}{\gminM}{\textbf{#1}}}
\newcommand{\gaugebarGT}[1]{\makegaugeG{#1}{\gmaxT}{\gminT}{#1}}
\newcommand{\gaugebarboldGT}[1]{\makegaugeG{#1}{\gmaxT}{\gminT}{\textbf{#1}}}
\newcommand{\gaugebarGD}[1]{\makegaugeG{#1}{\gmaxD}{\gminD}{#1}}
\newcommand{\gaugebarboldGD}[1]{\makegaugeG{#1}{\gmaxD}{\gminD}{\textbf{#1}}}

\newcommand{\gaugebarPM}[1]{\makegaugeP{#1}{\pmaxM}{\pminM}{#1}}
\newcommand{\gaugebarboldPM}[1]{\makegaugeP{#1}{\pmaxM}{\pminM}{\textbf{#1}}}
\newcommand{\gaugebarPT}[1]{\makegaugeP{#1}{\pmaxT}{\pminT}{#1}}
\newcommand{\gaugebarboldPT}[1]{\makegaugeP{#1}{\pmaxT}{\pminT}{\textbf{#1}}}
\newcommand{\gaugebarPD}[1]{\makegaugeP{#1}{\pmaxD}{\pminD}{#1}}
\newcommand{\gaugebarboldPD}[1]{\makegaugeP{#1}{\pmaxD}{\pminD}{\textbf{#1}}}

\begin{table*}[t]
\centering
\sisetup{detect-weight}
\setlength{\tabcolsep}{2pt}
\begin{tabular}{l r S[table-format=1.3] S[table-format=1.3] r r S[table-format=1.3] S[table-format=1.3] r r S[table-format=1.3] S[table-format=1.3] r}
\toprule
\multicolumn{1}{c}{Dataset} & \multicolumn{4}{c}{Mip-NeRF 360} & \multicolumn{4}{c}{Tanks\&Temples} & \multicolumn{4}{c}{Deep Blending} \\
\cmidrule(lr){0-0} \cmidrule(lr){2-5} \cmidrule(lr){6-9} \cmidrule(lr){10-13}
\multicolumn{1}{c}{Method}  & \multicolumn{1}{c}{PSNR$\uparrow$} & \multicolumn{1}{c}{SSIM$\uparrow$} & \multicolumn{1}{c}{LPIPS$\downarrow$} & \multicolumn{1}{c}{\#G/M$\downarrow$} & \multicolumn{1}{c}{PSNR$\uparrow$} & \multicolumn{1}{c}{SSIM$\uparrow$} & \multicolumn{1}{c}{LPIPS$\downarrow$} & \multicolumn{1}{c}{\#G/M$\downarrow$} & \multicolumn{1}{c}{PSNR$\uparrow$} & \multicolumn{1}{c}{SSIM$\uparrow$} & \multicolumn{1}{c}{LPIPS$\downarrow$} & \multicolumn{1}{c}{\#G/M$\downarrow$} \\
\addlinespace[2pt]
\midrule
\addlinespace[2pt]
3DGS \cite{kerbl2023} & \gaugebarPM{27.50} & 0.813 & 0.221 & \gaugebarGM{3.111} & \gaugebarPT{23.63} & 0.850 & \textbf{0.180} & \gaugebarGT{1.830} & \gaugebarPD{29.42} & 0.900 & 0.250 & \gaugebarGD{2.780} \\
LP-3DGS-R \cite{10.5555/3737916.3741807} & \gaugebarPM{27.47} & 0.812 & 0.227 & \gaugebarGM{1.959} & \gaugebarPT{23.60} & 0.842 & 0.188 & \gaugebarGT{1.244} & {--} & {--} & {--} & {--} \\
LP-3DGS-M \cite{10.5555/3737916.3741807} & \gaugebarPM{27.12} & 0.805 & 0.239 & \gaugebarGM{1.866} & \gaugebarPT{23.41} & 0.834 & 0.198 & \gaugebarGT{1.116} & {--} & {--} & {--} & {--} \\
Compact3DGS \cite{lee2023compact} & \gaugebarPM{27.08} & 0.798 & 0.247 & \gaugebarGM{1.388} & \gaugebarPT{23.32} & 0.831 & 0.201 & \gaugebarGT{0.836} & \gaugebarPD{29.79} & 0.901 & 0.258 & \gaugebarGD{1.060} \\
EAGLES \cite{girish2023eagles} & \gaugebarPM{27.23} & 0.809 & 0.238 & \gaugebarGM{1.330} & \gaugebarPT{23.37} & 0.840 & 0.200 & \gaugebarGT{0.650} & \gaugebarPD{29.86} & 0.910 & 0.250 & \gaugebarGD{1.190} \\
CompGS \cite{navaneet2023compact3d} & \gaugebarPM{27.12} & 0.806 & 0.240 & \gaugebarGM{0.845} & \gaugebarPT{23.44} & 0.838 & 0.198 & \gaugebarGT{0.520} & \gaugebarPD{29.90} & 0.907 & 0.251 & \gaugebarGD{0.550} \\
Mini-Splatting \cite{10.1007/978-3-031-72980-5_10} & \gaugebarPM{27.40} & 0.821 & 0.219 & \gaugebarGM{0.559} & \gaugebarPT{23.45} & 0.841 & 0.186 & \gaugebarGT{0.319} & \gaugebarPD{30.05} & 0.909 & 0.254 & \gaugebarGD{0.397} \\
LightGaussian* \cite{fan2023lightgaussian}& \gaugebarPM{26.81} & {0.788} & {0.273} & \gaugebarGM{0.451} & \gaugebarPT{23.10} & {0.804} & {0.228} & \gaugebarGT{0.251} & \gaugebarPD{27.57} & {0.824} & {0.298} & \gaugebarGD{0.302} \\
MaskGaussian* \cite{11094772} & \gaugebarPM{26.95} & {0.802} & {0.252} & \gaugebarGM{0.430} & \gaugebarPT{23.57} & {0.837} & {0.200} & \gaugebarGT{0.384} & \gaugebarPD{29.80} & {0.905} & {0.257} & \gaugebarGD{0.342} \\
\midrule
\addlinespace[2pt]
GaussianSpa \cite{11092501} & \gaugebarPM{27.25} & {0.812} & {0.225} & \gaugebarGM{0.426} & \gaugebarPT{23.40} & {0.841} & {0.188} & \gaugebarGT{0.295} & \gaugebarPD{29.98} & {0.910} & {0.252} & \gaugebarGD{0.262} \\
\textbf{Proposed (ours)} & \gaugebarPM{27.36} & {0.816} & {0.223} & \gaugebarGM{0.401} & \gaugebarPT{23.64} & {0.844} & {0.186} & \gaugebarGT{0.252} & \gaugebarPD{29.95} & {0.909} & {0.252} & \gaugebarGD{0.275} \\
\addlinespace[2pt]
\midrule
\addlinespace[2pt]
GaussianSpa* \cite{11092501}& \gaugebarPM{27.43} & {0.821} & {0.219} & \gaugebarGM{0.353} & \gaugebarPT{23.61} & {0.848} & {0.183} & \gaugebarGT{0.185} & \gaugebarPD{30.11} & {0.911} & {0.250} & \gaugebarGD{0.216} \\
\textbf{Proposed (ours)*} & \gaugebarboldPM{27.51} & \textbf{0.822} & \textbf{0.218} & \gaugebarboldGM{0.330} & \gaugebarboldPT{23.75} & \textbf{0.850} & {0.181} & \gaugebarboldGT{0.180} & \gaugebarboldPD{30.18} & \textbf{0.912} & \textbf{0.249} & \gaugebarboldGD{0.208} \\
\addlinespace[2pt]
\bottomrule
\end{tabular}
\caption{\textbf{Quantitative results on multiple datasets, compared with 3DGS pruning models (On-training time pruning).} * denotes 40,000 training iterations, used to enable a direct comparison with the full GaussianSpa configuration. 
\#G/M represents the number of Gaussians in millions.}
\label{tab:tab2}
\end{table*}


Fig.~\ref{fig:fig3} shows the distribution of the cumulative $\Delta_{SE}$ values, computed across all training views, on a log scale. The histogram clearly illustrates a long-tailed distribution, with the majority of Gaussians contributing to error values close to zero when being removed and very few leading to high error values. This suggests the existence of numerous redundant Gaussians that contribute little ($\Delta_{SE} \approx 0$) to the final rendered image. This distribution is a key motivation for our proposed pruning strategy, as it implies that we can identify and remove a large portion of these Gaussians while minimizing visual quality degradation, i.e., by removing those that would lead to very low error values.

\subsection{Simplification Framework}
\label{sec:sec3.4}
The GaussianPOP framework leverages its efficient error metric in a three-stage process: quantification, pruning, and fine-tuning. First, the quantification stage executes our quantification Algorithm~\ref{alg:error_quant} over the complete set of training views. This computes the cumulative $\Delta SE_k$ for every Gaussian in the scene, populating a global error buffer, $\Delta_{SE}$. Then, the pruning stage sorts all Gaussians globally by their accumulated $\Delta_{SE}$ score. Then, a target percentage $P$\% of the Gaussians with the lowest error score is pruned. This on-the-fly operation does not rely on backward gradients, unlike mask-based or sparsity losses. Finally, the fine-tuning stage allows the model to compensate for the removed Gaussians. The quantification strategy adapts to the specific application scenario. The on-training application integrates our metric as a pruning schedule applied at discrete steps during training. In contrast, post-training application applies a more accurate iterative re-quantification, an approach analyzed in detail in Sec. \ref{sec:sec4.3}.

\section{Experiments and Results}
We present a comprehensive evaluation to validate GaussianPOP. First, we detail our experimental setup (Sec. \ref{sec:sec4.1}). Second, we benchmark our method against state-of-the-art simplification techniques in both on-training and post-training pruning scenarios (Sec. \ref{sec:sec4.2}). Then, we provide ablation studies on our iterative quantification strategy (Sec. \ref{sec:sec4.3}), and analyze the computational complexity (Sec. \ref{sec:sec4.4}). We evaluate our method against two distinct categories of 3DGS simplification methods: on-training and post-training pruning methods. Despite being a criterion-based pruning method, our framework offers high flexibility while outperforming existing state-of-the-art methods in both scenarios.

\subsection{Experimental Conditions}
\label{sec:sec4.1}
\begin{figure*}[t]
 \centering 
 \includegraphics[width=\linewidth]{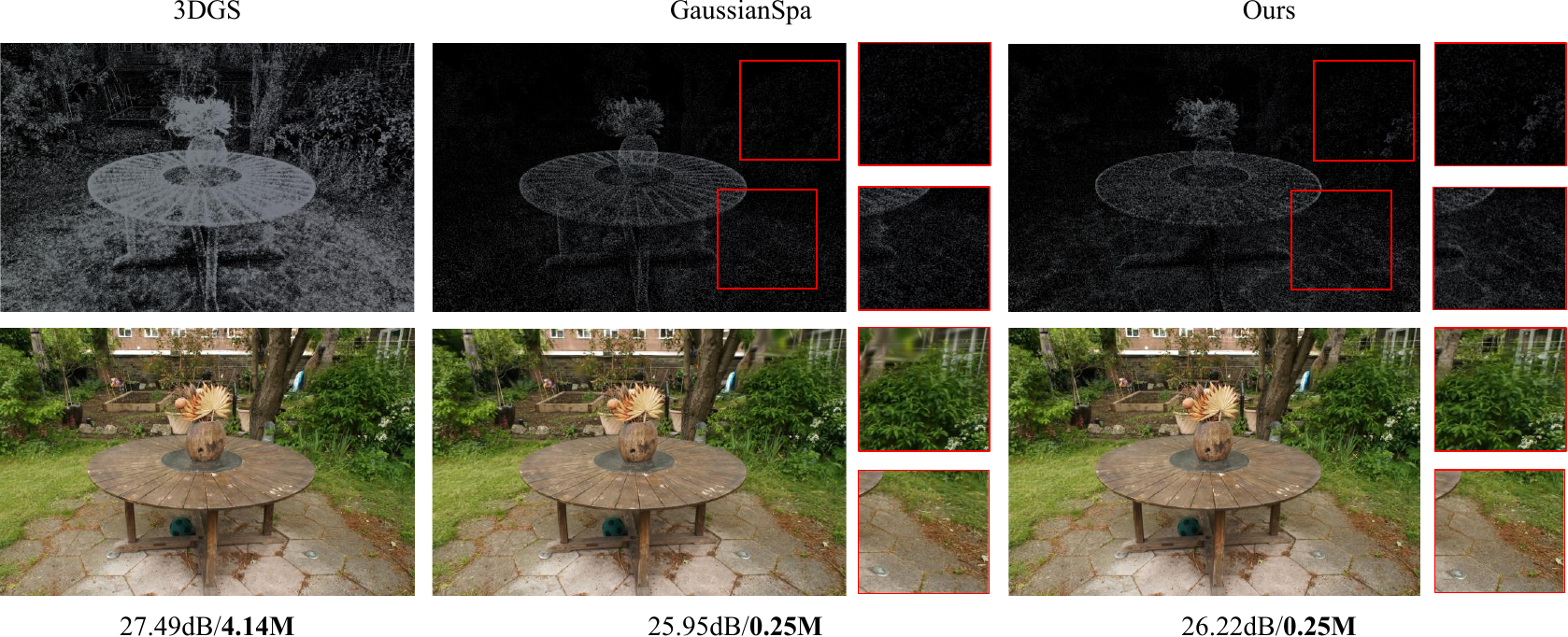}
 \caption{\textbf{Qualitative comparison for `Garden' scene at 0.25M Gaussians.} The density visualization (top row) illustrates our method achieves a more effective distribution for capturing detailed regions compared to GaussianSpa. This superior distribution results in a final render (bottom row) that preserves finer detail and texture, achieving a higher PSNR.}
 \label{fig:fig4}
\end{figure*}

\begin{figure}[t]
 \centering
          \begin{subfigure}[b]{0.49\linewidth}
         \centering
         \includegraphics[width=\linewidth]{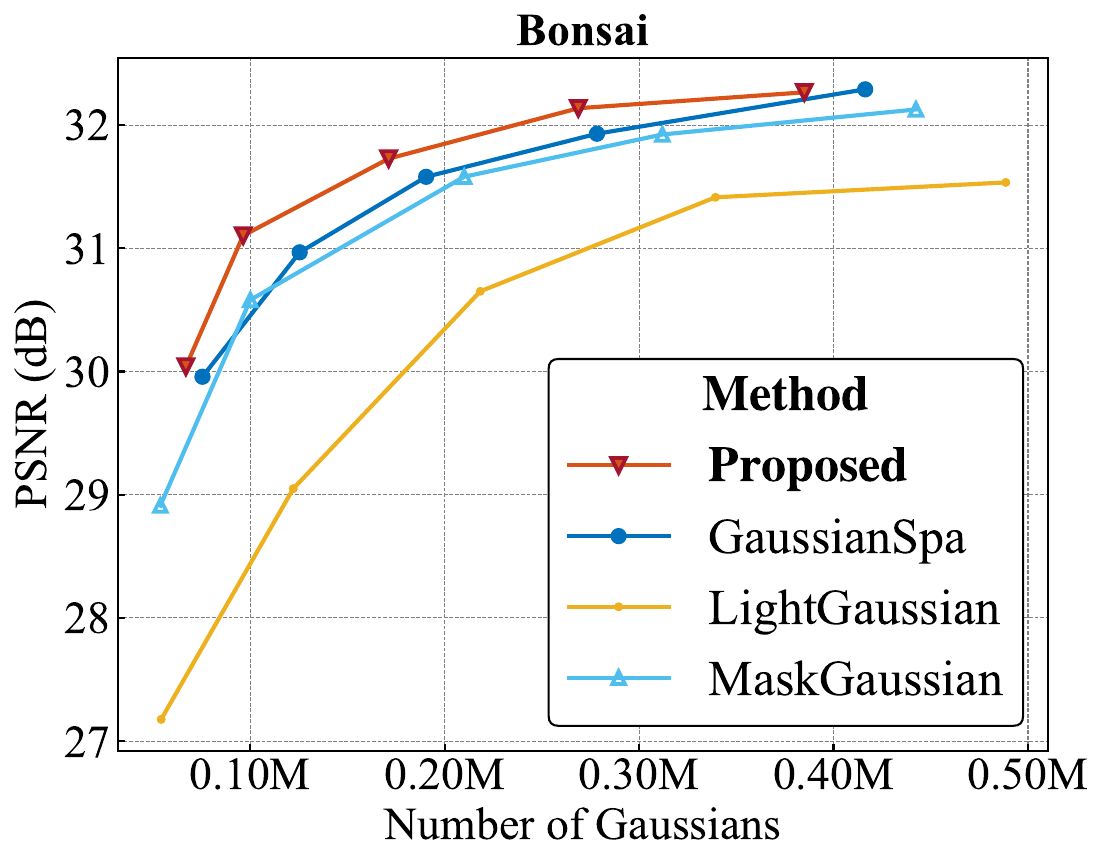}
     \end{subfigure}
          \begin{subfigure}[b]{0.49\linewidth}
         \centering
         \includegraphics[width=\linewidth]{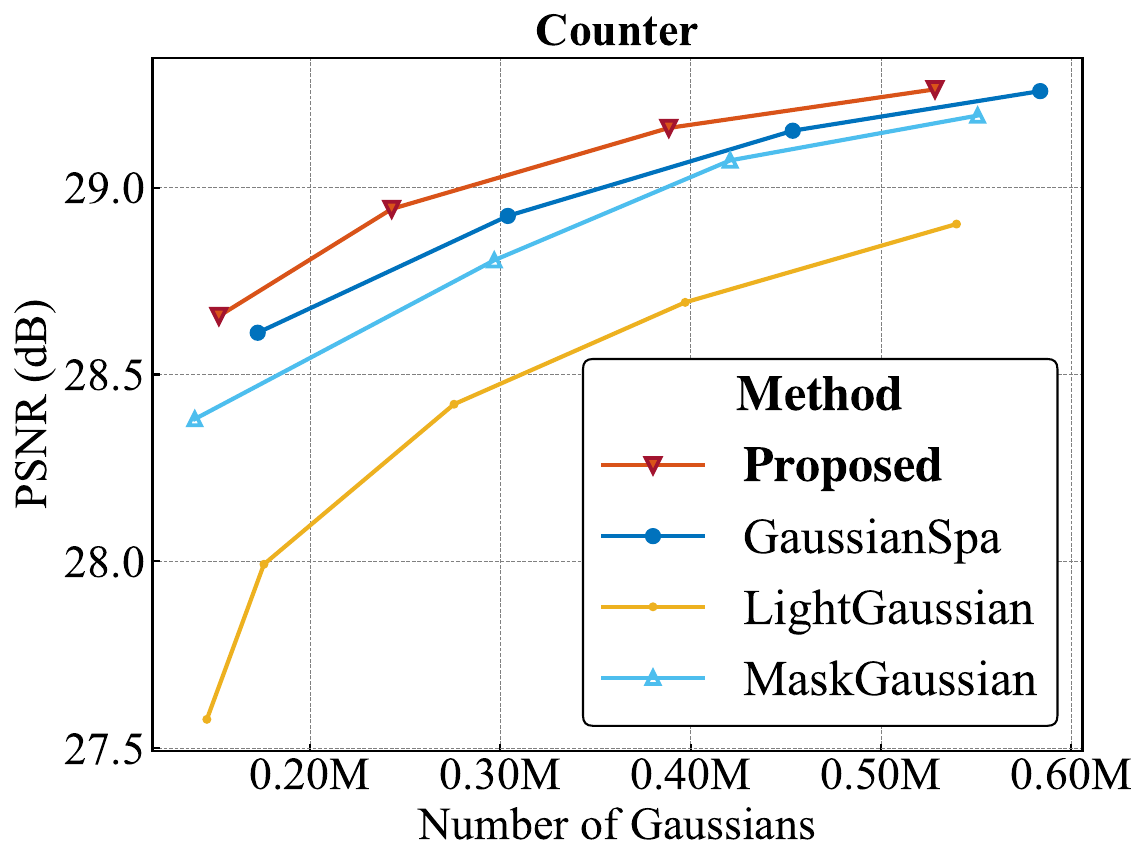}
     \end{subfigure}
 \caption{\textbf{Quality-compactness trade-off for on-training pruning.} The plots compare PSNR against the number of Gaussians for our method and competing approaches on two representative scenes.
 }
 \label{fig:fig5}
\end{figure}

\textbf{On-training Pruning Conditions.} We integrate our framework directly into the training process from scratch. To ensure a fair comparison, we followed the configuration from GaussianSpa, including its mini-splatting densification \cite{10.1007/978-3-031-72980-5_10}. Our pruning mechanism is applied twice during training: at 15k and 20k iterations for a 30k total training iteration, and at 20k and 25k iterations for a 40k total training iteration. At these points, our error criterion $\Delta_{SE}$ is computed, and a percentage of the Gaussians with the lowest error score is permanently removed. The training then continues, allowing the remaining Gaussians to optimize their parameters and compensate for the removal.

\textbf{Post-training Pruning Conditions.} We start from a fully pre-trained 3DGS model with 30k iterations. We evaluate the pruned model's performance immediately after pruning (post-prune) and after an optional 5k iterations (fine-tuning). For these experiments, we apply our criterion using 8 iterative cycles, as described in Sec. \ref{sec:sec4.3}. One advantage enabling the post-prune scenario is that our error quantification does not require ground truth views, like other criterion-based methods. This differs from sparsity induction models \cite{11094772,11092501}, which is integrated into the training loop and require continuous access to ground truth views for their loss calculation and the subsequent, mandatory fine-tuning process.

\begin{figure}[t]
 \centering
      \begin{subfigure}[b]{0.98\linewidth}
         \centering
         \includegraphics[width=\linewidth]{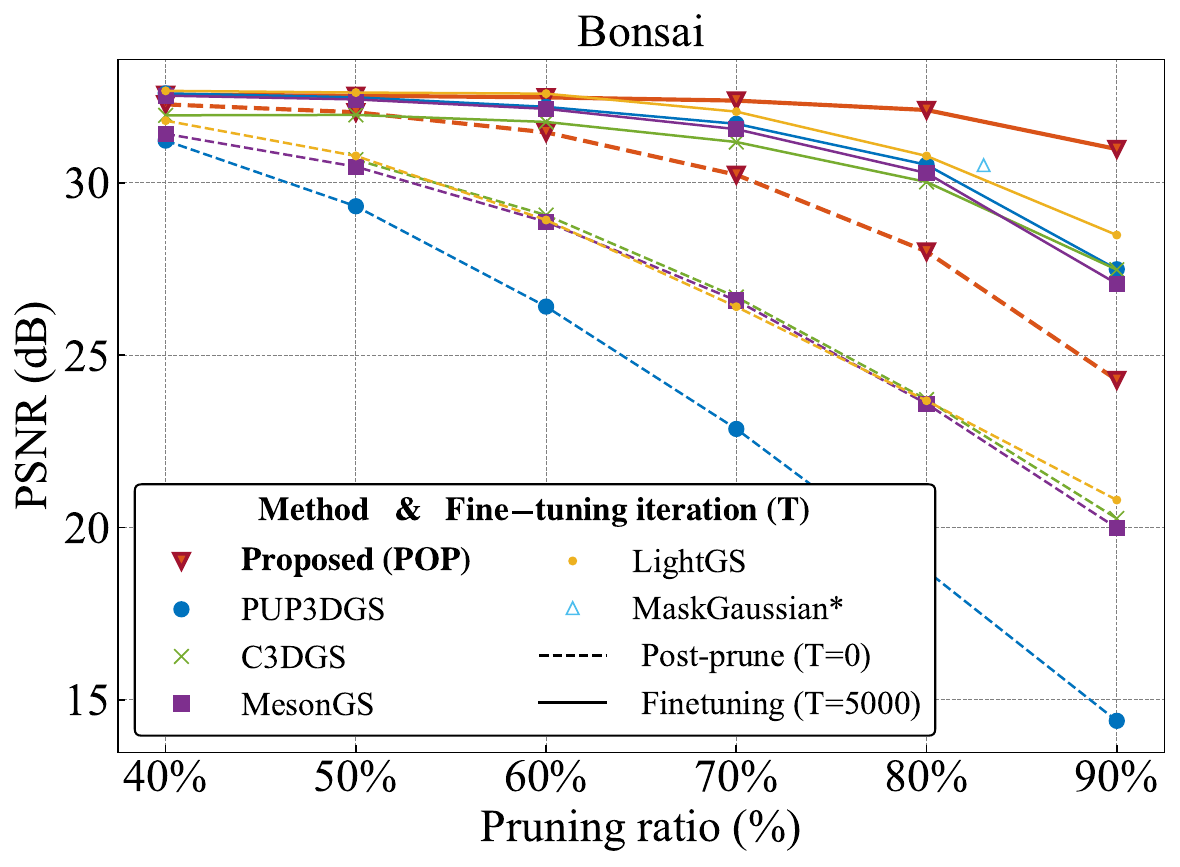}
         
     \end{subfigure}
     \vspace{0.01\linewidth}
          \begin{subfigure}[b]{0.98\linewidth}
         \centering
         \includegraphics[width=\linewidth]{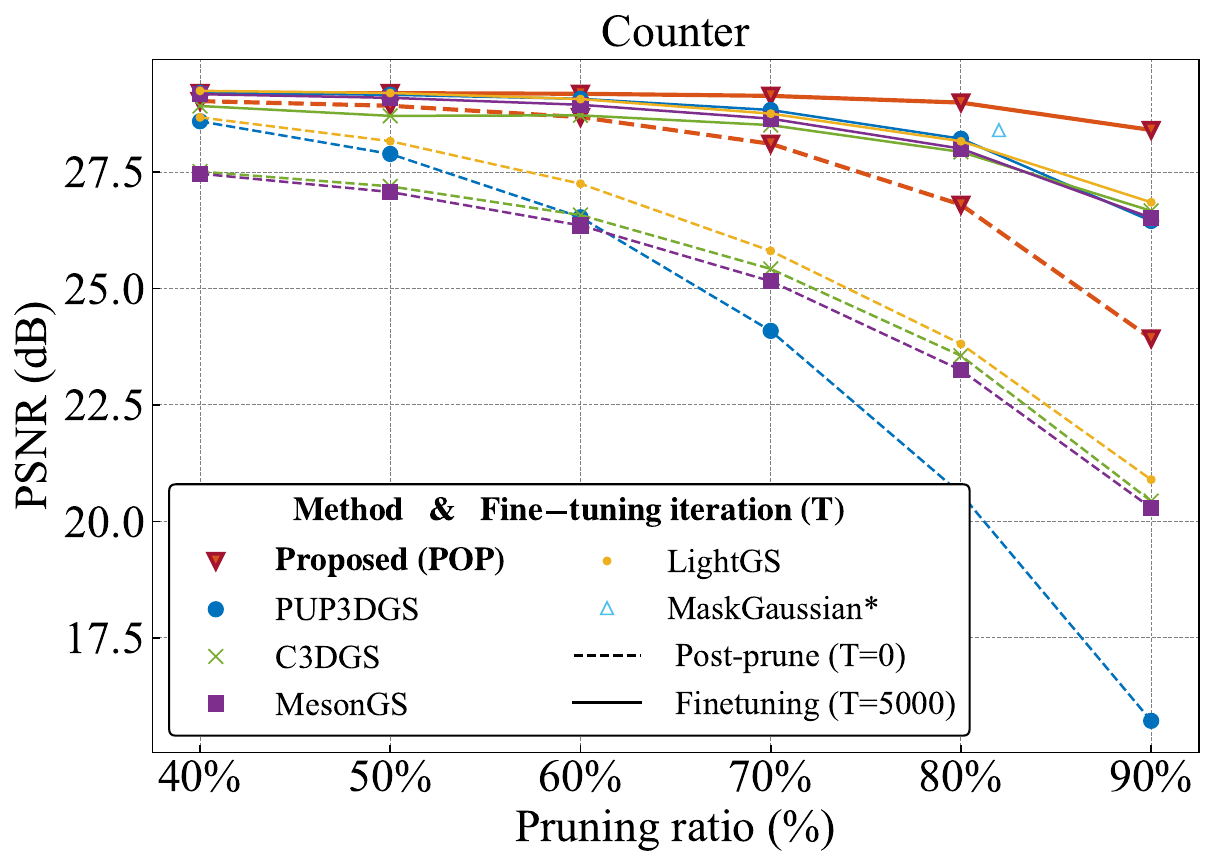}
        
     \end{subfigure}
         \caption{\textbf{Post-training pruning performance on the `Bonsai' and `Counter' scenes.} We compare our method against various heuristic-based criteria across a range of pruning ratios. 
         }
 \label{fig:fig6}
\end{figure}

\subsection{Experimental Results}
\label{sec:sec4.2}
We evaluate our method as both an on-training time pruning schedule and a post-training simplification tool. The proposed method consistently outperforms leading methods from both approaches, achieving a superior trade-off between model compactness and rendering quality.

\textbf{On-training Time Pruning Evaluations.} Despite being a criterion-based pruning method, our method consistently outperforms sparsity induction approaches, as shown in Tab. \ref{tab:tab2}. In the 30k iteration configuration, our method already demonstrates superior efficiency by achieving better rendering quality with a more compact model. This performance advantage is still clear in the 40k iteration configuration for full GaussianSpa model, where our model achieves a higher PSNR with a substantially smaller Gaussian count compared to the equivalent 40k GaussianSpa model. This consistent performance across all benchmarks validates our principled error quantification as a more effective simplification strategy, one that more accurately identifies and removes visually redundant information.

Fig. \ref{fig:fig4} provides a visual confirmation of this quantitative advantage. When comparing the models pruned to an identical 0.25M Gaussian count, the render from GaussianSpa exhibits a noticeable loss of texture and detail. Our method, at the same Gaussian count, produces a visibly sharper and more stable visual result. This provides direct visual evidence that our error metric is more effective at distinguishing visually critical details from redundant information than the heuristics used by competing methods. Fig. \ref{fig:fig5} shows the quality–compactness trade-off, where our method consistently defines a superior convex hull, achieving a higher PSNR for any given Gaussian budget. This result validates the quantitative findings, visually demonstrating the efficiency of our pruning criterion over competing approaches.

\textbf{Post-training Time Pruning Evaluations.} 
To specifically analyze the effectiveness of our $\Delta SE_{k}$ error criterion, we evaluate our method's performance as a post-training tool. First, we compare our method against other pruning criteria \cite{11092397,niedermayr2023compressed, fan2023lightgaussian}  across a range of pruning ratios on the `Bonsai' and `Counter' scenes. The results, in Fig.~\ref{fig:fig6}, show that even without fine-tuning (dashed lines), our method maintains a significantly higher PSNR than all competitors, especially at high pruning ratios. This validates that our metric is inherently more accurate at identifying non-essential Gaussians. 

\begin{figure}[t]
 \centering
          \begin{subfigure}[b]{0.49\linewidth}
         \centering
         \includegraphics[width=\linewidth]{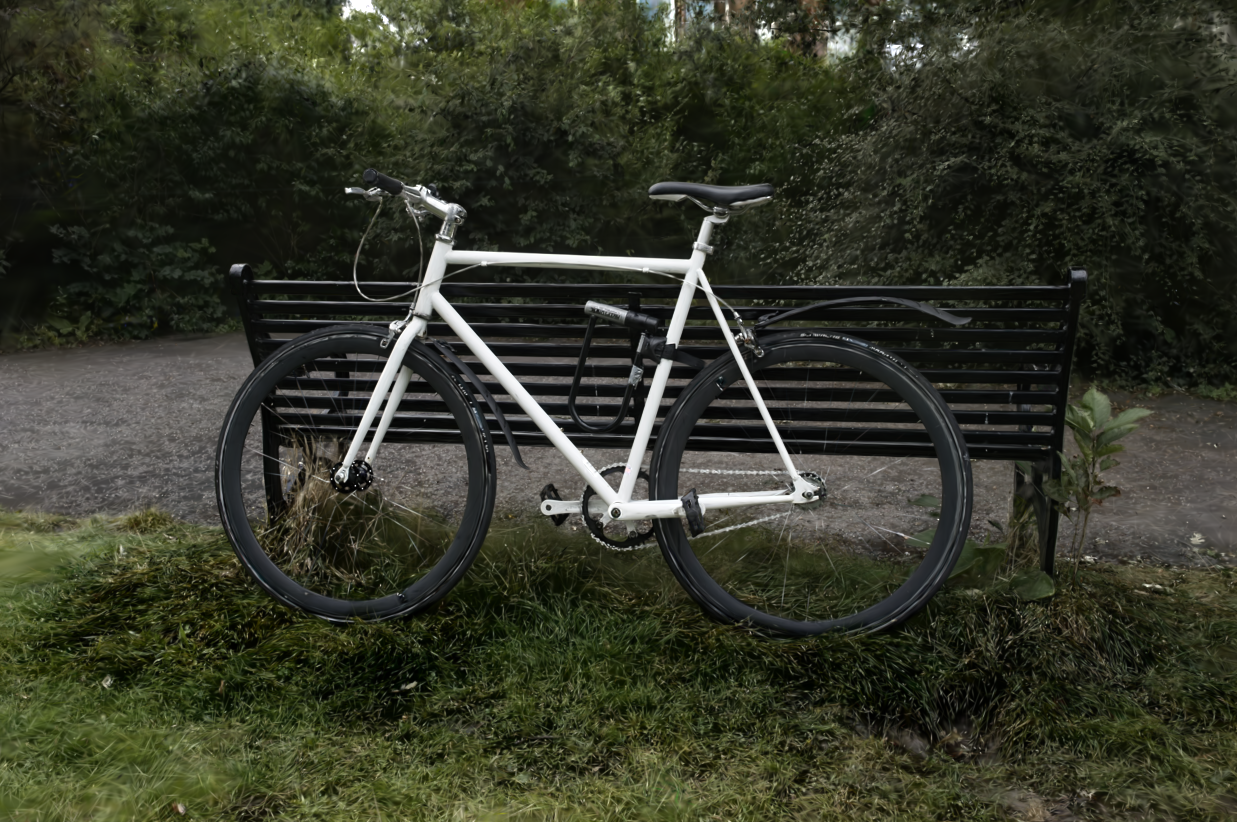}
         \subcaption{LightGaussian (80\% Pruned)}
     \end{subfigure}
          \begin{subfigure}[b]{0.49\linewidth}
         \centering
         \includegraphics[width=\linewidth]{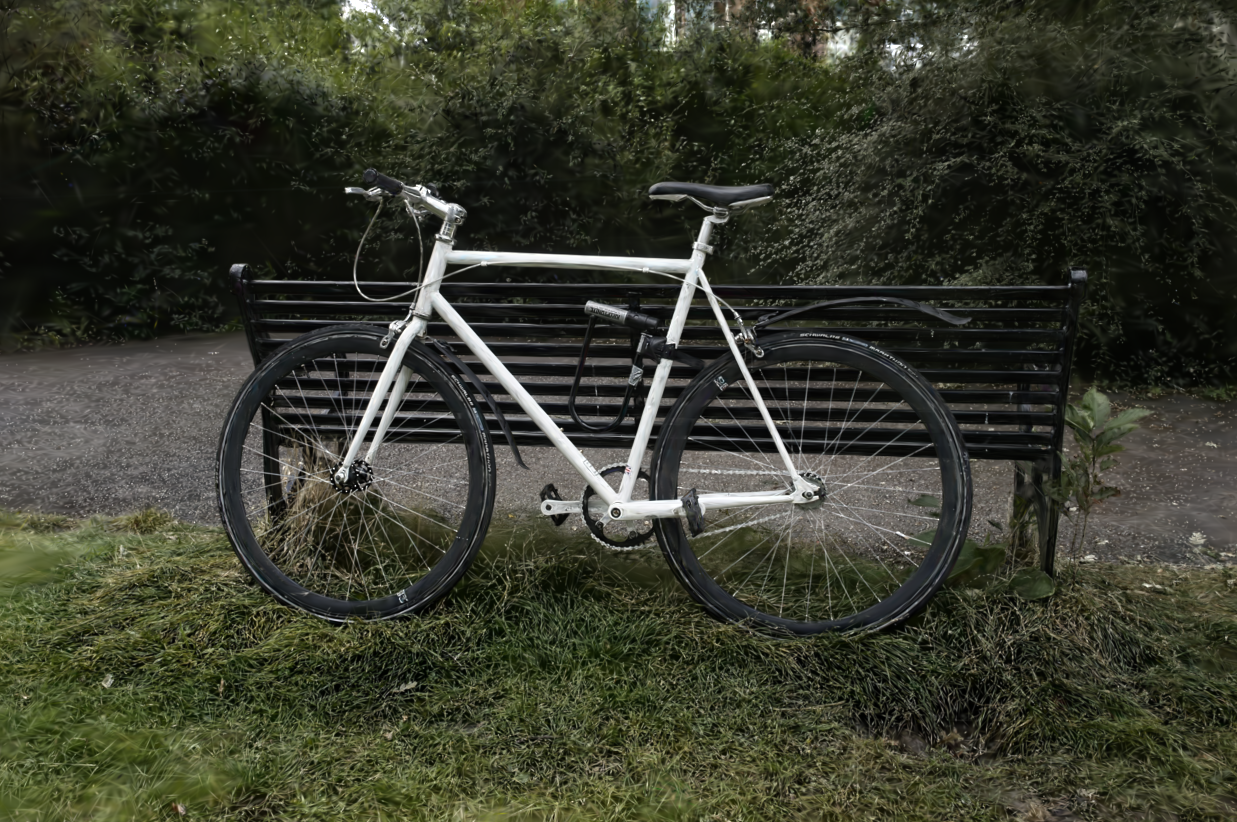}
         \subcaption{Ours (80\% Pruned)}
     \end{subfigure}
 \caption{\textbf{Qualitative comparison of post-prune results on the `Bicycle' scene.} Both models are pruned by 80\% from a pre-trained 3DGS model. Our method preserves fine details (see bicycle spokes) more effectively than LightGaussian \cite{fan2023lightgaussian}.}
 \label{fig:fig7}
\end{figure}

When fine-tuning (solid lines, 5k iterations) is applied, our method's performance shows remarkable stability and recovery, demonstrating that our criterion preserves a robust subset of Gaussians that adapt effectively. We also conducted a direct qualitative and quantitative comparison on the `Bicycle' scene, pruning 80\% of Gaussians from a pre-trained 3DGS model. As seen in Fig. \ref{fig:fig7}, our method preserves fine details more effectively than the pruning criterion of LightGaussian \cite{fan2023lightgaussian}, demonstrating a distinct advantage in reconstructing thin, high-frequency structures. As shown in Tab. \ref{tab:tab3}, with 80\% pruning and post-prune (T=0), our method achieves a PSNR of 20.71dB, outperforming LightGaussian's 18.46dB. After a 5k fine-tuning (+FT), our method maintains its lead, demonstrating the robustness of our $\Delta SE_{k}$ criterion.

\begin{table}[t]
\centering
\resizebox{\columnwidth}{!}{
\begin{tabular}{l|r@{\,$\rightarrow$\,}lr@{\,$\rightarrow$\,}lr@{\,$\rightarrow$\,}lc}
\toprule
\textbf{Method} & \multicolumn{2}{c}{\textbf{PSNR$\rightarrow$(+FT)}} & \multicolumn{2}{c}{\textbf{SSIM$\rightarrow$(+FT)}} & \multicolumn{2}{c}{\textbf{LPIPS$\rightarrow$(+FT)}} & \textbf{\#G/M} \\
\midrule
LightGS \cite{fan2023lightgaussian}& 18.46 & 24.35 & 0.537 & 0.713 & 0.364 & 0.272 & 0.97  \\
C3DGS \cite{niedermayr2023compressed} & 19.15 & 24.39 & 0.565 & 0.719 & 0.352 & 0.267 & 0.97  \\
PUP3DGS \cite{11092397}& 18.01 & 24.37 & 0.538 & 0.714 & 0.366 & 0.264 & 0.97  \\
MaskGaussian* \cite{11094772} & \multicolumn{1}{c}{N/A} & 24.88 & \multicolumn{1}{c}{N/A} & 0.739 & \multicolumn{1}{c}{N/A} & 0.255 & 1.02 \\
GaussianSpa* \cite{11092501} & \multicolumn{1}{c}{N/A} & 24.79 & \multicolumn{1}{c}{N/A} & 0.740 & \multicolumn{1}{c}{N/A} & 0.257 & 0.97 \\
\textbf{Ours (POP)} & \textbf{20.71} & \textbf{25.11} & \textbf{0.616} & \textbf{0.747} & \textbf{0.332} & \textbf{0.251} & 0.97\\ 
\bottomrule
\end{tabular}
}
\caption{\textbf{Quantitative comparison of post-training pruning methods on the `Bicycle' scene.} We report performance without fine-tuning (Post-prune) and with a 5,000 iteration fine-tuning (+FT) from the pruned model. }
\label{tab:tab3}
\end{table}

It is important to note that sparsity induction methods (denoted *), such as GaussianSpa \cite{11092501} and MaskGaussian \cite{11094772}, are excluded from the evaluations in the post-prune scenario. Because their simplification process is inherently tied to joint training, they do not provide a standalone pruning metric that can be applied to a pre-trained model. While those methods rely on joint training with sparsity loss to perform pruning every 1000 iterations, our method achieves superior performance even without modifying the 3DGS training process.

\begin{table}
  \centering
  \resizebox{\columnwidth}{!}{
  \begin{tabular}{c |c| c |c |S[table-format=1.2]}
    \toprule 
    \textbf{Scene} & \textbf{\begin{tabular}[c]{@{}c@{}}Re-quantifying \\ Cycle\end{tabular}} & \textbf{\begin{tabular}[c]{@{}c@{}}Total $\Delta_{SE}$ \\Threshold \end{tabular}} & \textbf{\begin{tabular}[c]{@{}c@{}}Actual PSNR Error (dB) \\ Training views (Test views)\end{tabular}} & {\textbf{\#G/M}} \\
    \midrule 
    \multirow{4}{*}{Bonsai}    
    & $C=1$ & \multirow{4}{*}{1500} & -2.98 (-1.69)    & 0.46     \\
    & $C=2$ &                       & -2.13 (-1.13)     & 0.47 \\
    & $C=4$ &                       & -1.56 (-0.83)     & 0.48 \\
    & $C=8$ &                       & -1.14 (-0.66)     & 0.48 \\ 
    \midrule 
    \multirow{4}{*}{Playroom}    
    & $C=1$ & \multirow{4}{*}{2500} & -2.53 (-0.78)     & 0.79     \\
    & $C=2$ &                       &  -1.94 (-0.60)    & 0.77     \\
    & $C=4$ &                       &  -1.57 (-0.48)    & 0.77 \\
    & $C=8$ &                       & -1.39 (-0.43)     & 0.77 \\ 
    \midrule 
    \multirow{4}{*}{Kitchen}     
    & $C=1$ & \multirow{4}{*}{2500} & -2.30 (-1.35)     & 0.75     \\
    & $C=2$ &                       &  -1.84 (-1.01)    & 0.76     \\
    & $C=4$ &                       &  -1.53 (-0.86)    & 0.76 \\
    & $C=8$ &                       &  -1.34 (-0.78)     & 0.77 \\ 
    \midrule 
    \multirow{4}{*}{Stump}       
    & $C=1$ & \multirow{4}{*}{5000} & -3.06 (-0.53)    & 1.61     \\
    & $C=2$ &                       & -2.26 (-0.45)     & 1.67 \\
    & $C=4$ &                       & -1.90 (-0.39)     & 1.74 \\
    & $C=8$ &                       & -1.75 (-0.36)     & 1.78 \\
    \bottomrule 
  \end{tabular}
  } 
  \caption{\textbf{Analysis of the effects of iterative re-quantification.} We compare the impact of increasing the re-evaluation cycle $C$, which significantly reduces the actual PSNR error, validating the robustness of the iterative approach.}
  \label{tab:tab4}
\end{table}

\subsection{Iterative Error Quantification}
\label{sec:sec4.3}
Post-training simplification has practical value for real-world deployment as it enables the immediate compaction of pre-trained models without the prohibitive computational overhead of re-training from scratch. This allows users to efficiently optimize existing legacy 3DGS models for resource-constrained devices. However, post-training simplification applies pruning to a static, fully-trained model, typically involving the removal of a large percentage of Gaussians all at once. This difference necessitates a more precise error quantification to avoid significant quality degradation. 

Our $\Delta SE_k$ criterion, being a direct measure of visual error, enables budget-based pruning. We set an absolute total error threshold $B$ and remove all Gaussians whose $\Delta SE_k$ falls below it. However, a naive one-shot calculation of this budget is a greedy approximation. The challenge lies in the $\alpha$-blending process, where removing a large batch of Gaussians simultaneously affects the transmittance $T_j$ for all subsequent Gaussians $j>k$. This cascading effect is not captured by the single-pass $\Delta SE_{k}$ calculation. Consequently, this one-shot quantification tends to underestimate the errors, leading to a cumulative visual impact that is larger than the expected error budget. This highlights the necessity of our iterative re-quantification approach.

We analyze an iterative re-quantification approach, as shown in Tab.~\ref{tab:tab4}. Negative PSNR error denotes PSNR decrease relative to the unpruned baseline. Instead of applying the total budget $B$ at once, we divide it across $C$ iterative cycles, applying only a partial budget $B/C$ per cycle. Crucially, we re-calculate the error $\Delta SE_k$ for all remaining Gaussians at the start of each new cycle. This iterative cycle ensures error metrics are updated to reflect visual changes from previous removals, leading to a much more accurate and stable simplification. The results demonstrate the clear superiority of this iterative approach. A single-pass removal results in the largest actual PSNR error, highlighting the one-shot limitation. In contrast, by re-evaluating the $\Delta SE_k$ values after redundant low $\Delta SE_k$ Gaussians are removed, the error calculation is more accurate than a single-pass quantification. Consequently, this approach achieves a more accurate pruning, significantly reducing the actual PSNR estimation error and better aligning the actual training view error with the intended error budget.

\subsection{Complexity Analysis}
\label{sec:sec4.4}
We analyze the computational cost of our quantification stage, benchmarked on an NVIDIA A100 GPU, with results detailed in Tab.~\ref{tab:tab5}. In the on-training scenario, our method can reduce the total training time compared to the original 3DGS model for 30k training. This acceleration occurs because pruning significantly reduces the Gaussian count, which accelerates all subsequent training iterations. The overhead of the quantification itself is negligible, requiring only 10.52s for two quantification passes. 

In the post-training scenario, we applied different iterative cycles, and the total cost increases with the number of cycles. However, the cost per iteration progressively decreases, from 4.76s on the first pass to 3.83s. This demonstrates the substantial quality gains clearly justify the minimal overhead, as the model achieves 90\% compactness from the original pre-trained model (23.62dB/1.812M) with only a 0.1dB PSNR drop after 5k fine-tuning.

\begin{table}[t]
  \centering
  
  \subcaptionbox{\textbf{On-training time complexity for `Bicycle' scene.}  A comparison of the total training time of our method with two pruning steps against the 3DGS baseline.}{%
    \resizebox{\columnwidth}{!}{ 
      \begin{tabular}{c | c c c} 
      \hline
      \textbf{Method} & \textbf{Pruning Iteration} & \textbf{Total Training Time} & \textbf{\#G/M} \\ \hline
      3DGS (Bicycle) & N/A & 27m11s & 4.87 \\
      Ours (30k), $P=[0.5, 0.5]$ & [15000, 20000] & \textbf{24m49s} & 1.23 \\
      GaussianSpa (30k), $\kappa=[0.5, 0.5]$ & [15000, 20000] & 25m33s & 1.23 \\
      MaskGaussian (30k), $\lambda_m=0.1$ & Every 1000 & 26m21s & 1.45 \\
      \hline
      \end{tabular}
    }
  } 

  \vspace{2.5mm} 
  
 \subcaptionbox{\textbf{Post-training iterative re-evaluation for `Tanks\&Temples' dataset.} Analysis of computational cost and visual quality for different re-evaluation cycles. \label{tab:post_training}}{%
   \resizebox{\columnwidth}{!}{ 
     \begin{tabular}{c | c c c c} 
     \hline
     \textbf{Method} & \textbf{Quantification Time Cost} & \textbf{PSNR$\rightarrow$(+FT)} & \textbf{\#G/M} \\ \hline
     $C=1, P=0.9$  & 4.76s/iter (Total 4.76s) &  17.50dB$\rightarrow$23.12dB & 0.18 \\
     $C=2, P=0.9$  &  4.45s/iter (Total 8.89s)  & 18.92dB$\rightarrow$23.35dB & 0.18 \\
     $C=4, P=0.9$  & 4.01s/iter (Total 16.0s) & 19.53dB$\rightarrow$23.43dB & 0.18 \\
     $C=8, P=0.9$  & 3.83s/iter (Total 30.6s) & 19.62dB$\rightarrow$23.54dB & 0.18 \\ \hline
     \end{tabular}
   }
 } 

  \caption{\textbf{Analysis of computational complexity for each scenario.} (a) On-training time scenario and (b) Post-training iterative re-evaluation scenario.}
  \label{tab:tab5} 
\end{table}


\section{Conclusion}
\label{sec:conclusion}
We presented GaussianPOP, a principled simplification framework that analytically quantifies the visual error of each Gaussian directly from the rendering equation. Our method introduces an efficient single-pass algorithm that makes this quantification practical with minimal overhead. The error-based criterion supports both on-training pruning and post-training simplification via iterative re-quantification, which further improves stability and reduces cumulative visual degradation. Consequently, GaussianPOP provides a robust and general method for compact representation, showing a superior trade-off between model compactness and rendering fidelity. These results underscore the advantages of direct error quantification over conventional importance scores for 3DGS simplification.

\clearpage
{
    \small
    \bibliographystyle{ieeenat_fullname}
    \bibliography{main}
}


\maketitlesupplementary

\section{Derivation of the Background Term}
To formally justify our error criterion, we define the background term $b_{k+1}$ as the color of the scene occluded by the first $k$ Gaussians. First, we split the rendering equation into the accumulated color up to the $k$-th Gaussian $P_k$ and the contribution of the remaining Gaussians $j > k$,

$$C_{render} = \sum_{i=1}^{N} T_i \alpha_i c_i = \underbrace{\sum_{i=1}^{k} T_i \alpha_i c_i}_{P_k} + \sum_{j=k+1}^{N} T_j \alpha_j c_j$$The second term represents the attenuated background contribution. Since the transmittance $T_j$ for any $j > k$ inherently includes the accumulated transmittance $T_{k+1}$, we factor $T_{k+1}$ out to define the background term,

$$\sum_{j=k+1}^{N} T_j \alpha_j c_j = T_{k+1} \underbrace{\sum_{j=k+1}^{N} \frac{T_j}{T_{k+1}} \alpha_j c_j}_{b_{k+1}}$$Physically, $b_{k+1}$ represents the unattenuated background color directly behind the $k$-th Gaussian. Normalizing by $T_{k+1}$ removes the foreground occlusion, enabling a meaningful comparison between the Gaussian's color $c_k$ and the background $b_{k+1}$. Thus, $\Delta SE_k$ serves as a measure of visual redundancy, identifying primitives that can be pruned with minimal color loss due to their high similarity to the underlying background. Here, the term inside the summation defines $b_{k+1}$. Substituting this back into the decomposition yields,

$$C_{render} = P_k + T_{k+1} b_{k+1}$$We rearrange this equation to calculate $b_{k+1}$ using the final rendered color and accumulated values, avoiding the need for re-computation in our algorithm,

$$b_{k+1} = \frac{C_{render} - P_k}{T_{k+1}}$$

\section{Implementation Details}
In our implementation, we fixed the maximum number of contributing Gaussians per pixel ($N_{max}$) to 64. We conducted experiments increasing $N_{max}$ to 128 and observed no meaningful difference in the final pruning results or rendering quality. For comparison, we employ a two-stage pruning with $P=0.75$ for Mip-NeRF 360 outdoor scenes and $P=\{0.65, 0.7\}$ for other datasets depending on number of points. To benchmark GaussianSpa at a comparable scale, we modified the `pruning ratio2' from the reference per-scene configuration.

We evaluated the Mip-NeRF 360 dataset using the pre-downsampled images (\lstinline{images_2} and \lstinline{images_4}) instead of the automatic downsampling. Due to different downsampling algorithms, these pre-downsampled images preserve fine details, making the reconstruction task considerably more difficult. This protocol yields lower PSNR (e.g., 25.6 to 25.2dB for bicycle) for all comparisons but provides a more accurate evaluation of detail preservation.

\section{Per-scene Quantitative Results} 
%

\begin{table}[b]
    \centering
    \resizebox{\columnwidth}{!}{
    \begin{tabular}{llcccc}
        \toprule
        \multicolumn{6}{c}{\textbf{Tanks \& Temples}} \\
        \midrule
        Scene & Method & PSNR$\uparrow$ & SSIM$\uparrow$ & LPIPS$\downarrow$ & \#G/M$\downarrow$ \\
        \midrule
        \multirow{4}{*}{Train} 
            & 3DGS & 21.94 & 0.815 & 0.210 & 1.110 \\
            & Mini-Splatting & 21.78 & 0.805 & 0.231 & 0.287 \\
            & GaussianSpa & 21.97 & 0.815 & 0.228 & 0.200 \\
            & Ours & 22.21 & 0.815 & 0.224 & 0.177 \\ 
        \midrule
        \multirow{4}{*}{Truck} 
            & 3DGS & 25.31 & 0.885 & 0.150 & 2.540 \\
            & Mini-Splatting & 25.13 & 0.878 & 0.141 & 0.352 \\
            & GaussianSpa & 25.25 & 0.881 & 0.138 & 0.170 \\ 
            & Ours & 25.28 & 0.885 & 0.138 & 0.183 \\ 
        \midrule
        \midrule
        \multirow{4}{*}{\textbf{Average}} 
            & 3DGS & 23.63 & 0.850 & \textbf{0.180} & 1.825 \\
            & Mini-Splatting & 23.46 & 0.841 & 0.186 & 0.320 \\
            & GaussianSpa & 23.61 & 0.848 & 0.183 & 0.185 \\ 
            & \textbf{Ours} & \textbf{23.75} & \textbf{0.850} & 0.181 & \textbf{0.180} \\ 
        \bottomrule
    \end{tabular}
    }
    
    \centering
    \resizebox{\columnwidth}{!}{
    \begin{tabular}{llcccc}
        \toprule
        \multicolumn{6}{c}{\textbf{Deep Blending}} \\
        \midrule
        Scene & Method & PSNR$\uparrow$ & SSIM$\uparrow$ & LPIPS$\downarrow$ & \#G/M$\downarrow$ \\
        \midrule
        \multirow{4}{*}{DrJohnson} 
            & 3DGS & 28.77 & 0.900 & 0.250 & 3.260 \\
            & Mini-Splatting & 29.37 & 0.904 & 0.261 & 0.377 \\
            & GaussianSpa & 29.51 & 0.909 & 0.247 & 0.214 \\
            & Ours & 29.62 & 0.911 & 0.246 & 0.195 \\ 
        \midrule
        \multirow{4}{*}{Playroom} 
            & 3DGS & 30.07 & 0.900 & 0.250 & 2.290 \\
            & Mini-Splatting & 30.72 & 0.914 & 0.248 & 0.417 \\
            & GaussianSpa & 30.71 & 0.913 & 0.253 & 0.218 \\
            & Ours & 30.74 & 0.913 & 0.252 & 0.221 \\ 
        \midrule
        \midrule
        \multirow{4}{*}{\textbf{Average}} 
            & 3DGS & 29.42 & 0.900 & 0.250 & 2.775 \\
            & Mini-Splatting & 30.05 & 0.909 & 0.254 & 0.397 \\
            & GaussianSpa & 30.11 & 0.911 & 0.250 & 0.216 \\ 
            & \textbf{Ours} & \textbf{30.18} & \textbf{0.912} & \textbf{0.249} & \textbf{0.208} \\ 
        \bottomrule
    \end{tabular}
    }
    \caption{\textbf{`On-training pruning', Quantitative comparison on Tanks \& Temples and Deep Blending datasets.} Note that both GaussianSpa and our method are trained for 40k iterations.}
    \label{tab:tt_deepblending_full}
\end{table}


\begin{table}[b]
    \centering
    \caption{\textbf{`On-training pruning', Quantitative comparison on Mip-NeRF 360 dataset.} We compare our method with 3DGS, Mini-Splatting, and GaussianSpa. Note that both GaussianSpa and our method are trained for 40k iterations.}
    \resizebox{\columnwidth}{!}{
    \begin{tabular}{llcccc}
        \toprule
        Scene & Method & PSNR$\uparrow$ & SSIM$\uparrow$ & LPIPS$\downarrow$ & \#G/M$\downarrow$ \\
        \midrule
        \multirow{4}{*}{Bicycle} 
            & 3DGS & 25.13 & 0.750 & 0.240 & 5.310 \\
            & Mini-Splatting & 25.21 & 0.760 & 0.247 & 0.696 \\
            & GaussianSpa & 25.26 & 0.758 & 0.257 & 0.465 \\
            & Ours & 25.08 & 0.755 & 0.260 & 0.395 \\
        \midrule
        \multirow{4}{*}{Bonsai} 
            & 3DGS & 32.19 & 0.950 & 0.180 & 1.250 \\
            & Mini-Splatting & 31.73 & 0.945 & 0.180 & 0.360 \\
            & GaussianSpa & 31.92 & 0.945 & 0.180 & 0.287 \\
            & Ours & 32.05 & 0.947 & 0.178 & 0.265 \\
        \midrule
        \multirow{4}{*}{Counter} 
            & 3DGS & 29.11 & 0.910 & 0.180 & 1.170 \\
            & Mini-Splatting & 28.53 & 0.911 & 0.184 & 0.308 \\
            & GaussianSpa & 28.85 & 0.917 & 0.178 & 0.314 \\
            & Ours & 29.01 & 0.919 & 0.176 & 0.325 \\
        \midrule
        \multirow{4}{*}{Flowers} 
            & 3DGS & 21.37 & 0.590 & 0.360 & 3.470 \\
            & Mini-Splatting & 21.42 & 0.616 & 0.336 & 0.670 \\
            & GaussianSpa & 21.57 & 0.609 & 0.335 & 0.356 \\
            & Ours & 21.55 & 0.605 & 0.339 & 0.330 \\
        \midrule
        \multirow{4}{*}{Garden} 
            & 3DGS & 27.32 & 0.860 & 0.120 & 5.690 \\
            & Mini-Splatting & 26.99 & 0.842 & 0.156 & 0.788 \\
            & GaussianSpa & 26.72 & 0.838 & 0.159 & 0.377 \\
            & Ours & 26.81 & 0.840 & 0.158 & 0.345 \\
        \midrule
        \multirow{4}{*}{Kitchen} 
            & 3DGS & 31.53 & 0.930 & 0.120 & 1.770 \\
            & Mini-Splatting & 31.24 & 0.929 & 0.122 & 0.438 \\
            & GaussianSpa & 31.53 & 0.934 & 0.117 & 0.316 \\
            & Ours & 31.75 & 0.936 & 0.115 & 0.325 \\
        \midrule
        \multirow{4}{*}{Room} 
            & 3DGS & 31.59 & 0.920 & 0.200 & 1.500 \\
            & Mini-Splatting & 31.44 & 0.929 & 0.193 & 0.394 \\
            & GaussianSpa & 31.46 & 0.928 & 0.190 & 0.302 \\
            & Ours & 31.50 & 0.932 & 0.186 & 0.288 \\
        \midrule
        \multirow{4}{*}{Stump} 
            & 3DGS & 26.73 & 0.770 & 0.240 & 4.420 \\
            & Mini-Splatting & 27.35 & 0.803 & 0.219 & 0.717 \\
            & GaussianSpa & 26.88 & 0.806 & 0.222 & 0.377 \\
            & Ours & 27.04 & 0.808 & 0.218 & 0.365 \\
        \midrule
        \multirow{4}{*}{Treehill} 
            & 3DGS & 22.61 & 0.640 & 0.350 & 3.420 \\
            & Mini-Splatting & 22.69 & 0.652 & 0.332 & 0.663 \\
            & GaussianSpa & 22.71 & 0.655 & 0.335 & 0.387 \\
            & Ours & 22.79 & 0.655 & 0.335 & 0.335 \\
        \midrule
        \midrule
        \multirow{4}{*}{\textbf{Average}} 
            & 3DGS & 27.50 & 0.813 & 0.221 & 3.111 \\
            & Mini-Splatting & 27.40 & 0.821 & 0.219 & 0.559 \\
            & GaussianSpa & 27.43 & 0.821 & 0.219 & 0.353 \\
            & \textbf{Ours} & \textbf{27.51} & \textbf{0.822} & \textbf{0.218} & \textbf{0.330} \\
        \bottomrule
    \end{tabular}
    }
    \label{tab:mipnerf360_full}
\end{table}

\begin{table}[t]
    \centering
    \caption{\textbf{`Post-training pruning', Quantitative comparison on all datasets.} Fine-tuning is performed for 5,000 iterations with $C=8$.}
    \resizebox{\columnwidth}{!}{
    \begin{tabular}{llcccc}
        \toprule
        Scene & Threshold & PSNR$\uparrow$ & SSIM$\uparrow$ & LPIPS$\downarrow$ & \#G/M$\downarrow$ \\
        \midrule
        \multirow{3}{*}{Bicycle} 
             & $P=0.7$ & 25.17 & 0.751 & 0.249 & 1.460 \\
             & $P=0.8$ & 25.12 & 0.747 & 0.252 & 0.973 \\
             & $P=0.9$ & 24.32 & 0.672 & 0.350 & 0.487 \\
        \midrule
        \multirow{3}{*}{Bonsai} 
             & $P=0.7$ & 32.18 & 0.941 & 0.213 & 0.321 \\
             & $P=0.8$ & 32.09 & 0.936 & 0.223 & 0.214 \\
             & $P=0.9$ & 30.98 & 0.914 & 0.262 & 0.107 \\
        \midrule
        \multirow{3}{*}{Counter} 
             & $P=0.7$ & 29.14 & 0.908 & 0.207 & 0.324 \\
             & $P=0.8$ & 29.00 & 0.902 & 0.219 & 0.216 \\
             & $P=0.9$ & 28.40 & 0.875 & 0.263 & 0.108 \\
        \midrule
        \multirow{3}{*}{Flowers} 
             & $P=0.7$ & 21.58 & 0.602 & 0.354 & 0.876 \\
             & $P=0.8$ & 21.51 & 0.589 & 0.375 & 0.584 \\
             & $P=0.9$ & 21.08 & 0.542 & 0.428 & 0.292 \\
        \midrule
        \multirow{3}{*}{Garden} 
             & $P=0.7$ & 27.39 & 0.863 & 0.120 & 1.243 \\
             & $P=0.8$ & 27.15 & 0.849 & 0.144 & 0.828 \\
             & $P=0.9$ & 26.28 & 0.799 & 0.221 & 0.414 \\
        \midrule
        \multirow{3}{*}{Kitchen} 
             & $P=0.7$ & 31.57 & 0.927 & 0.131 & 0.480 \\
             & $P=0.8$ & 31.38 & 0.922 & 0.141 & 0.320 \\
             & $P=0.9$ & 30.56 & 0.902 & 0.179 & 0.160 \\
        \midrule
        \multirow{3}{*}{Room} 
             & $P=0.7$ & 31.72 & 0.921 & 0.224 & 0.392 \\
             & $P=0.8$ & 31.62 & 0.918 & 0.230 & 0.261 \\
             & $P=0.9$ & 31.26 & 0.907 & 0.259 & 0.131 \\
        \midrule
        \multirow{3}{*}{Stump} 
             & $P=0.7$ & 26.70 & 0.776 & 0.227 & 1.285 \\
             & $P=0.8$ & 26.73 & 0.776 & 0.233 & 0.857 \\
             & $P=0.9$ & 26.36 & 0.747 & 0.287 & 0.428 \\
        \midrule
        \multirow{3}{*}{Treehill} 
             & $P=0.7$ & 22.60 & 0.630 & 0.363 & 0.975 \\
             & $P=0.8$ & 22.53 & 0.604 & 0.413 & 0.650 \\
             & $P=0.9$ & 22.13 & 0.526 & 0.503 & 0.325 \\
        \midrule
        \midrule
        \multirow{3}{*}{Train} 
             & $P=0.7$ & 22.13 & 0.813 & 0.214 & 0.326 \\
             & $P=0.8$ & 22.06 & 0.802 & 0.237 & 0.218 \\
             & $P=0.9$ & 21.80 & 0.768 & 0.292 & 0.109 \\
        \midrule
        \multirow{3}{*}{Truck} 
             & $P=0.7$ & 25.39 & 0.884 & 0.152 & 0.616 \\
             & $P=0.8$ & 25.34 & 0.879 & 0.161 & 0.411 \\
             & $P=0.9$ & 25.16 & 0.859 & 0.205 & 0.205 \\
        \midrule
        \midrule
        \multirow{3}{*}{Playroom} 
             & $P=0.7$ & 30.18 & 0.909 & 0.251 & 0.554 \\
             & $P=0.8$ & 30.19 & 0.909 & 0.255 & 0.369 \\
             & $P=0.9$ & 30.11 & 0.906 & 0.270 & 0.185 \\
        \midrule
        \multirow{3}{*}{DrJohnson} 
             & $P=0.7$ & 29.46 & 0.906 & 0.243 & 0.938 \\
             & $P=0.8$ & 29.47 & 0.906 & 0.245 & 0.625 \\
             & $P=0.9$ & 29.38 & 0.901 & 0.261 & 0.313 \\
        \bottomrule
    \end{tabular}
    }
    \label{tab:mipnerf360_full}
\end{table}

\pagebreak

\begin{figure*}
    \centering
     \includegraphics[width=\linewidth]{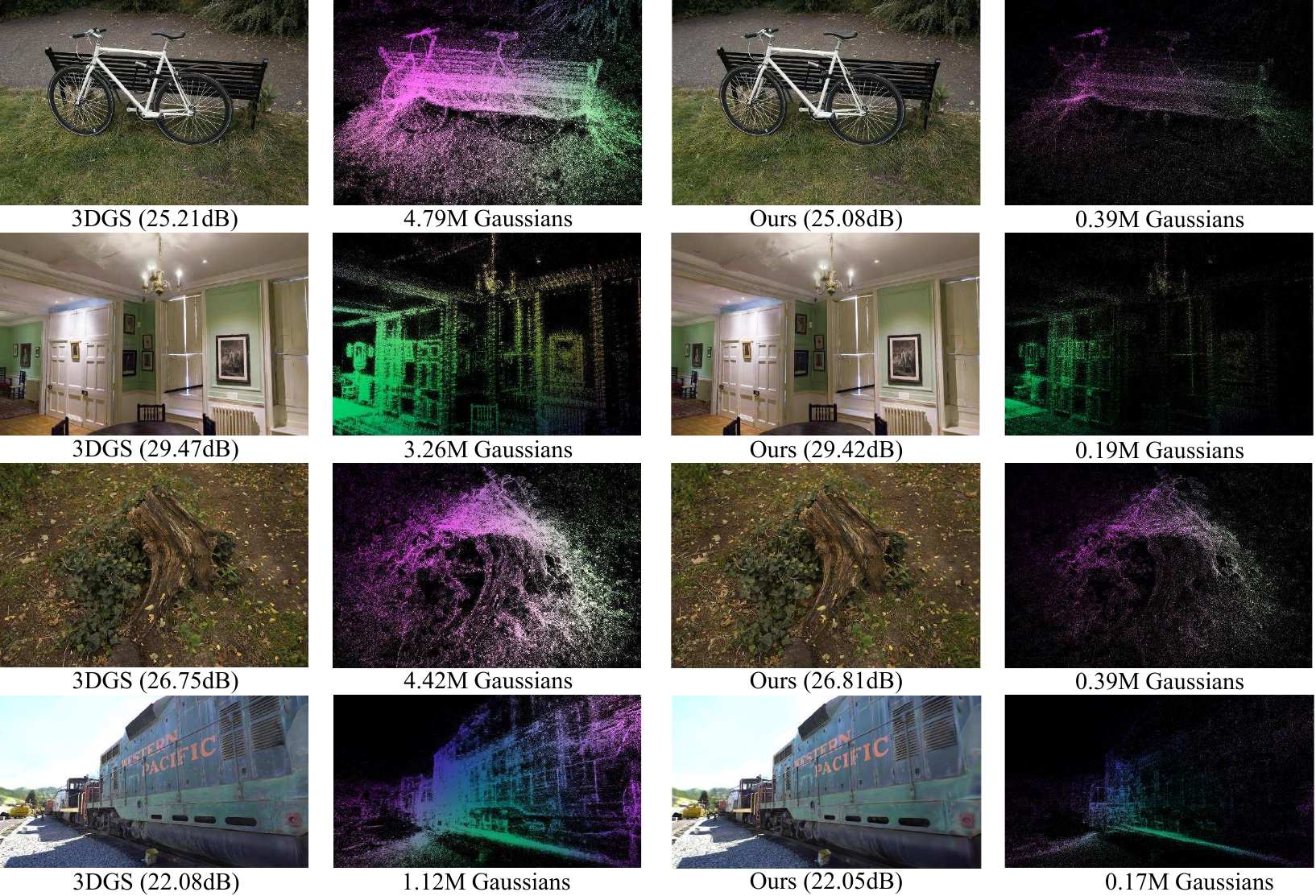}
     
     \caption{\textbf{Qualitative comparison of `on-training pruning' on representative scenes.} We compare our method with the 3DGS baseline across different datasets. Each subfigure reports the rendering quality (PSNR) and the number of Gaussians. Please zoom in for a detailed view.
     }
\label{sfig1}
\end{figure*}

\begin{figure*}
     \centering
     \includegraphics[width=\linewidth]{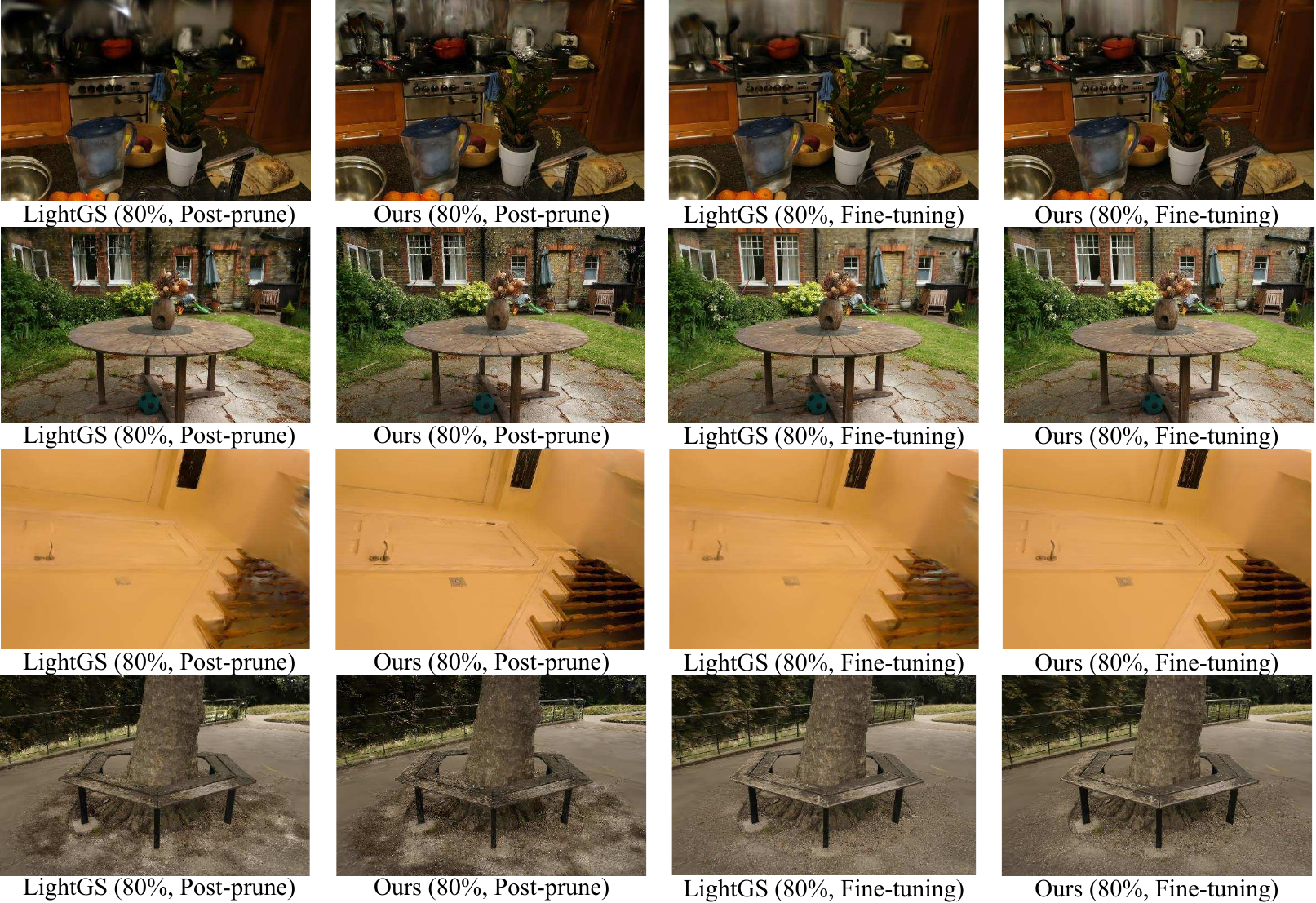}
     
     \caption{\textbf{Qualitative comparison of `post-training pruning' on representative scenes.} We compare our method against LightGS with an 80\% pruning ratio on selected benchmark scenes. `Post-prune' denotes results immediately after pruning, while `Fine-tuning' shows results after 5,000 refinement iterations. Please zoom in for a detailed view.
     }
\label{sfig2}
\end{figure*}

\end{document}